\pdfoutput=1

\documentclass[11pt]{article}

\usepackage[]{ACL2023}

\usepackage{times}
\usepackage{latexsym}
\usepackage{enumitem}
\usepackage{amsmath}
\usepackage{graphics}
\usepackage[]{graphicx}
\usepackage{subfig}
\usepackage{array}
\usepackage[T1]{fontenc}

\usepackage[utf8]{inputenc}

\usepackage{microtype}

\usepackage{inconsolata}

\title{I Spy a Metaphor: Large Language Models and Diffusion Models Co-Create Visual Metaphors}

\author{Tuhin Chakrabarty\textsuperscript{1}\thanks{*Equal contribution.}, ~~Arkadiy Saakyan\textsuperscript{1}\footnotemark[1], ~~Olivia Winn\textsuperscript{1}\footnotemark[1], ~~Artemis Panagopoulou\textsuperscript{2} \\ \textbf{Yue Yang\textsuperscript{2}, ~~Marianna Apidianaki\textsuperscript{2}, ~~Smaranda Muresan\textsuperscript{1}}
\\ \textsuperscript{1}Columbia University \hspace{8pt} \textsuperscript{2}University of Pennsylvania\\
\texttt{\{tuhin.chakr,a.saakyan,olivia\}@cs.columbia.edu} \\ 
}

\begin{document}
\maketitle
\begin{abstract}
Visual metaphors are powerful rhetorical devices used to persuade or communicate creative ideas through images.Similar to linguistic metaphors, they convey meaning implicitly through symbolism and juxtaposition of the symbols. We propose a new task of generating visual metaphors from linguistic metaphors. This is a challenging task for diffusion-based text-to-image models, such as DALL$\cdot$E 2, since it requires the ability to model implicit meaning and compositionality. We propose to solve the task through the collaboration between Large Language Models (LLMs) and Diffusion Models: Instruct GPT-3 (davinci-002) with Chain-of-Thought prompting generates text that represents a visual elaboration of the linguistic metaphor containing the implicit meaning and relevant objects, which is then used as input to the diffusion-based text-to-image models.Using a human-AI collaboration framework, where humans interact both with the LLM and the top-performing diffusion model, we create a high-quality dataset containing 6,476 visual metaphors for 1,540 linguistic metaphors and their associated visual elaborations. Evaluation by professional illustrators shows the promise of LLM-Diffusion Model collaboration for this task . To evaluate the utility of our Human-AI collaboration framework and the quality of our dataset, we perform both an intrinsic human-based evaluation and an extrinsic evaluation using visual entailment as a downstream task.
\end{abstract}

\section{Introduction}

\begin{figure}[!t]
\small
\centering
    \includegraphics[width=7.7cm]{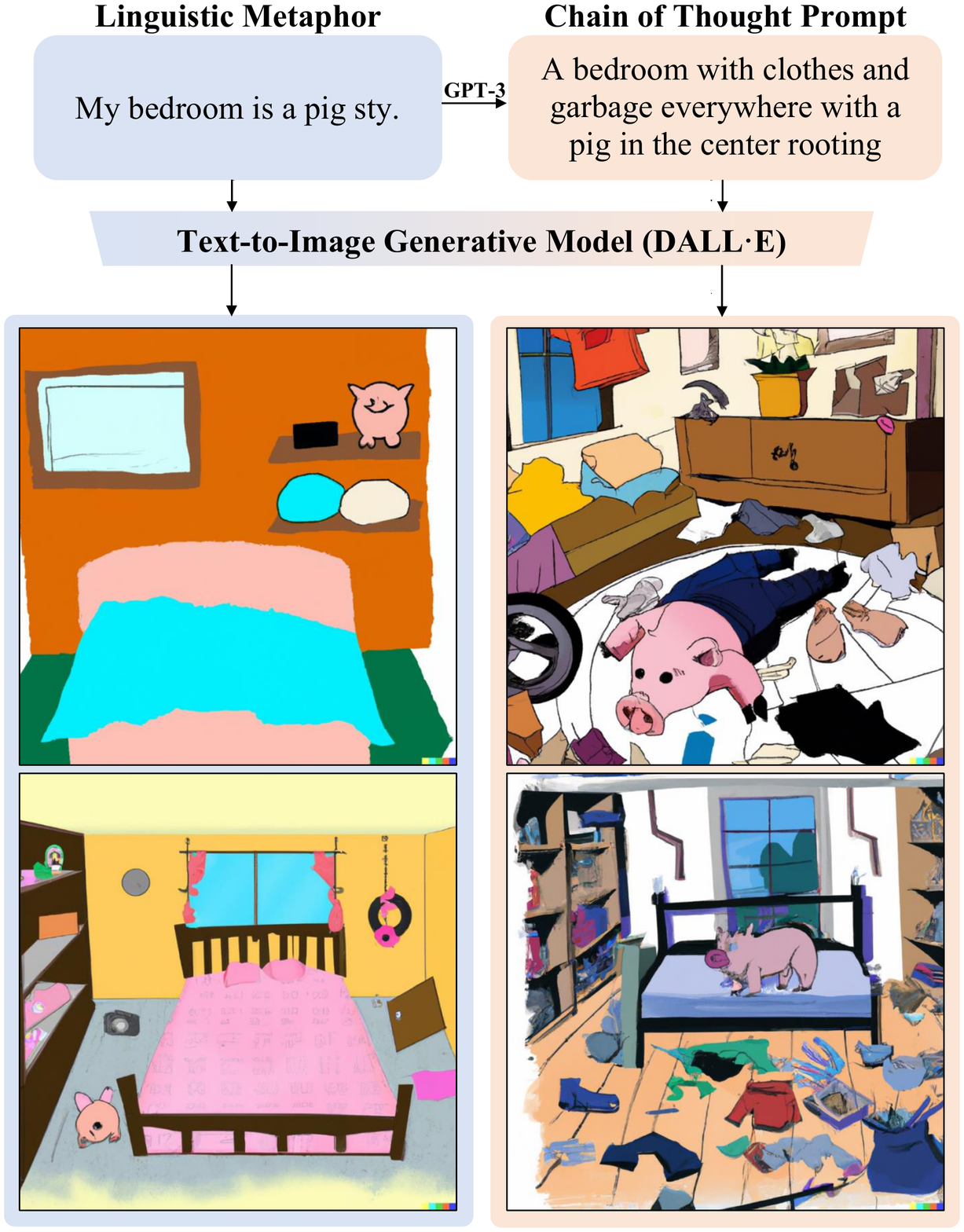}
    \caption{Visual metaphors generated by DALL$\cdot$E 2 for the linguistic metaphor ``My bedroom is a pig sty".We can take the original verbal metaphor as the input (left) or use GPT-3 with Chain of Thought prompting (right).}
    \label{fig:pigsty}
    \vspace{-3ex}
\end{figure}

Visual metaphors are rhetorical devices that serve to communicate a message through an image. They are often used as a means of persuasion in advertising \cite{phillips2003understanding,PhillipsandMcQuarrie:2004}, as their use leads to more favorable attitude toward the ad \cite{McquarrieandMick:1999}. Similarly to linguistic metaphors \cite{lakoff1993}, a visual metaphor takes a concept from a source domain and applies it to a target domain. In the case of visual metaphors, these domains need to be in some way visually grounded. 

Large diffusion-based text-to-image models, such as DALL$\cdot$E 2 \cite{ramesh2022hierarchical}, PARTI \cite{yu2022scaling}, Stable Diffusion \cite{Rombach_2022_CVPR}, or IMAGEN \cite{saharia2022photorealistic}, can generate visually compelling images conditioned on input texts. However, in order to generate visual metaphors from linguistic metaphors, models are required to first identify the implicit meaning as well as the objects, properties, and relations involved, and then find a way to combine them in the generated image. For instance, given the linguistic metaphor ``\textit{My bedroom is a pig sty}'', as shown in Figure \ref{fig:pigsty}, a model would ideally need to extract the implicit meaning of the bedroom being ``\textit{messy}'', and then compose the concepts ``\textit{Bedroom}'', ``\textit{Messy}'' \& ``\textit{Pig}''. However, as shown in the left two images, when presented just with the linguistic metaphor, DALL$\cdot$E 2 generates images of a bedroom where pink is the prevalent color (perhaps due to pig's skin color), sometimes with the presence of a pig as a toy in a corner, and with little indication of a mess in the room.

The visual metaphor generation task is greatly impacted by two common challenges in text-to-image models, namely \textit{under-specification} and  \textit{attribute-object binding} \cite{hutchinson2022underspecification, ramesh2022hierarchical,saharia2022photorealistic}. Under-specification refers to the fact that finite and reasonable-length linguistic descriptions of real-world scenes by necessity omit a great deal of visual information \cite{hutchinson2022underspecification}. Attribute Binding is the task of binding the attributes to the correct objects, and is a fundamental problem for a more complex and reliable compositional generalization. Our proposed contributions address these challenges: 
\begin{itemize}[leftmargin=*]
    \itemsep0em 
    \item \textbf{A novel approach for generating visual metaphors through the collaboration of large language models (LLMs) and diffusion-based text-to-image models.} 
     Our LLM --- Instruct GPT-3 (davinci-002) \cite{NEURIPS2022_b1efde53} with   \textbf{Chain-of-Thought} (CoT) prompting \cite{NEURIPS2022_9d560961} --- generates a {\bf  visual elaboration} 
    of the linguistic metaphors. To design our CoT prompting elements, we take inspiration from prior work on VisualBlends \cite{10.1145/3290605.3300402} that put an emphasis on the objects to be represented in the visual metaphor. In addition, we also consider the implicit meaning to finally generate a visual elaboration that contains the essential objects and the implicit meaning of the linguistic metaphor. For our linguistic metaphor ``\textit{My bedroom is a pig sty}'', the visual elaboration generated by Instruct GPT-3 (davinci-002) with CoT prompting is ``\textit{A bedroom with clothes and garbage everywhere with a pig in the center rooting around.}'' (See Table \ref{table:prompt}). The generated visual elaboration becomes the input to diffusion-based text-to-image models such as DALL$\cdot$E 2 or Stable Diffusion to generate visual metaphors (see Figure 1 right).
    \item {\bf A high-quality visual metaphor dataset built through Human-AI collaboration.} 
    We propose a collaboration between humans, LLM, and the top-performing diffusion-based model (DALL$\cdot$E 2) to create a high-quality dataset of \textbf{6,476} visually metaphoric images. These represent \textbf{1,540} distinct linguistic metaphors and their associated visual elaborations generated through CoT prompting. We call our dataset \textbf{\texttt{HAIVMet}} (\textbf{H}uman-\textbf{AI} \textbf{V}isual \textbf{Met}aphor) 
(Section 3).
    \item 
    {\bf A thorough  evaluation of LLM-Diffusion Model collaboration and Human-AI collaboration.} In order to evaluate the power of LLM-Diffusion Model collaboration, we recruit professional illustrators and designers and ask them to compare the output of DALL$\cdot$E 2 and Stable Diffusion v2.1 when the input corresponds to the linguistic metaphor alone, or to the LLM-produced visual elaboration. Our evaluation shows the power of the LLM-Diffusion Model collaboration and the superiority of DALL$\cdot$E 2 compared to Stable Diffusion v2.1 (Section \ref{models}). To evaluate the utility of Human-AI collaboration and the quality of our dataset, we perform an intrinsic evaluation using the same expert evaluators and an extrinsic evaluation using a downstream task (Section \ref{haieval}). For the latter, we choose the Visual Entailment task: given an image and a hypothesis sentence, the model is asked to predict whether the sentence is implied by the image. We show that fine-tuning a state-of-the-art vision-language model on our dataset leads to $\sim$23-points improvement in accuracy compared to when it is only finetuned on SNLI-VE  \cite{xie2019visual}, a large-scale visual entailment dataset. 
\end{itemize}

We release our dataset, code, prompts, and illustrator  annotations at \url{https://github.com/tuhinjubcse/VisualMetaphors}.

\section{Related Work}
\paragraph{Generative Art.}
There has recently been a huge surge of AI-generated artwork and imagery with the new diffusion-based models being substantially better than previous Variational Autoencoders (VAE) and Generative Adversarial Networks (GANs). Some of the most popular current models are DALL$\cdot$E 2 \cite{https://doi.org/10.48550/arxiv.2204.06125}, MidJourney,\footnote{\url{https://www.midjourney.com/}} Craiyon,\footnote{\url{https://www.craiyon.com/}} and Stable Diffusion 
\cite{rombach2021highresolution}. These image generation models are able to handle a wide variety of prompts, though recent work has shown that there are still aspects of accurate depiction that these models fail to capture \cite{https://doi.org/10.48550/arxiv.2210.12889}. Recently,  \citet{kleinlein2022language} showed that diffusion models can handle language that is content-based and aimed at a neutral description of the scene, and fail to capture the underlying abstraction of figurative language. Recent work has also explored cutting-edge systems showcasing the power of large language models and text-to-image models in aiding creative processes across various applications. \citet{wang2023reelframer} present PopBlends, a system that leverages traditional knowledge extraction methods and large language models to automatically generate conceptual blends for pop culture references, significantly increasing the number of blend suggestions while reducing mental demand for users. Similarly, \citet{liu2023generative} introduce Generative Disco, an AI system that generates music visualizations using large language models and text-to-image models, offering an enjoyable, expressive, and easy-to-use tool for professionals in the creative field. \citet{wang2023reelframer} present ReelFramer, a system where GPT4 and DALLE2 collaborate in order to assist journalists in transforming written news stories into engaging short video narratives, by generating scripts, character boards, and storyboards. The proposed user study shows ReelFramer's effectiveness in easing the process and making framing exploration rewarding for journalism students.

\paragraph{Visual Metaphor.}
Visual metaphors are often abstract and can be challenging to interpret. \citet{10.1145/3325480.3325503} test several theories about how people interpret visual metaphors. They find that visual metaphors are interpreted correctly, without explanatory text, with 41.3\% accuracy. \citet{IndurkhyaandOjha:2013} highlight  the important role of perceptual similarity between the source and the target image (in terms of color, shape, etc) in metaphor comprehension and  creative interpretation.

\citet{artemis2021} propose the ArtEmis dataset which contains emotion attribution and explanation annotations for 80K artworks from WikiArt, including several visual metaphors and similes. Their dataset serves to train captioning systems to express emotions and associated explanations derived from visual stimuli, instead of generating images conditioned on text. \citet{Zhang2021} collect a multimodal metaphor dataset from Twitter posts and advertisement posters that contain a metaphor in the caption, in the image, or both. However, they do not generate any new data and, as of yet, the data has not been publicly released.  
\citet{liu2022opal} release Opal, a system that guides users in generating diverse and relevant text-to-image illustrations for news articles by utilizing structured exploration. Unlike research on generating textual metaphors \cite{yu2019,chakrabarty2020,chakrabarty2021,veale2016,abe2006,terai2010}, visual metaphor generation has received less attention. \citet{https://doi.org/10.48550/arxiv.2212.09898} proposed MetaCLUE, a set of vision tasks that serve to evaluate the metaphor understanding and generation capabilities of state-of-the-art vision and language models. Their results show that most tested state-of-the-art models struggle to produce satisfactory results, in both a zero-shot and a finetuning setting. \citet{hwang2023memecap} focus on building a dataset for captioning and interpreting memes that are a widely popular tool for web users to express their thoughts using visual metaphors. More recently,  \citet{yosef2023irfl} present the Image Recognition of Figurative Language dataset, designed to evaluate vision and language models' understanding of figurative language, including metaphors, similes, and idioms. The dataset features multimodal examples and introduces two novel benchmark tasks, aimed at promoting the development of models that can effectively comprehend figurative language.Current baseline models have shown significantly poorer performance compared to human understanding, highlighting the challenges this domain poses for machine learning.

\section{Human-AI Collaboration for Visual Metaphor Dataset Creation}
\label{sec:data}

\begin{figure}[!t]
\small
\centering
    \includegraphics[width=8cm]{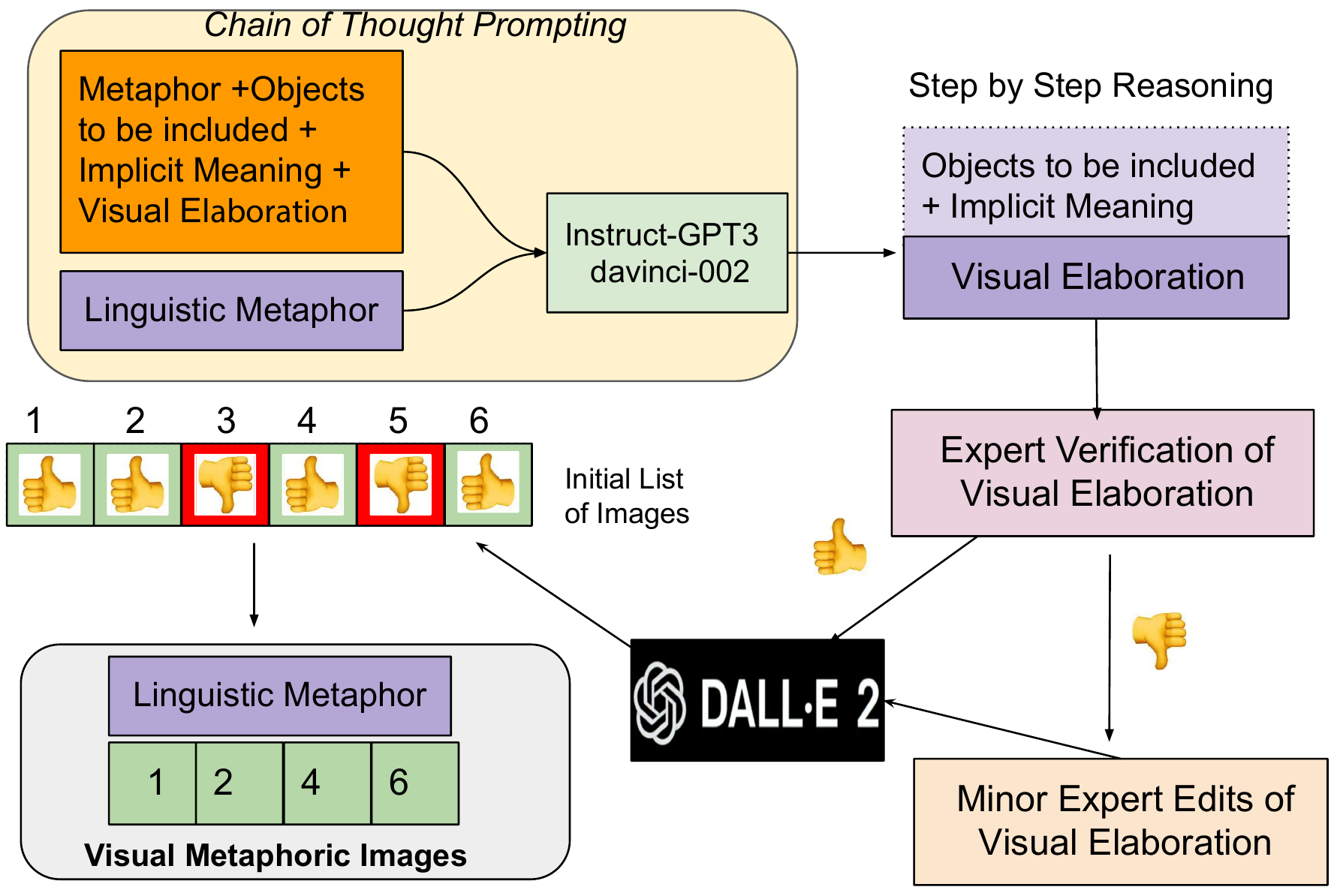}
    \caption{Human-AI collaboration framework (LLMs-Diffusion Model-Humans). Instruct GPT-3 with CoT prompting generates visual elaborations from linguistic metaphors, which are then validated and possibly edited by humans, if necessary. Visual elaborations are then used as input to DALL$\cdot$E 2 to generate visual metaphors. Experts filter poor-quality visual metaphors. For example, images 3 and 5 in the figure are discarded by experts.}
    \label{fig:pipeline}
\end{figure}

We propose a three-step Human-AI collaboration approach for generating visual metaphors from linguistic metaphors. This process involves 1) selecting linguistic metaphors that are visually grounded; 2) using large language models to generate visual elaborations of linguistic metaphors that capture relevant objects and implicit meaning, with expert edits when required; 3) using diffusion-based models to generate visual metaphors from visual elaborations, with filtering of low quality samples by experts. A detailed pipeline diagram for our dataset creation is shown in Figure \ref{fig:pipeline}.

We source our linguistic metaphors from six resources, removing any duplicates:
\textbf{\texttt{FLUTE}} \cite{chakrabarty2022flute},
\textbf{\texttt{Advertisements}} \cite{Hussain2017AutomaticUO},
\textbf{\texttt{CoPoet}} \cite{chakrabarty2022help},
\textbf{\texttt{FigQA}} \cite{https://doi.org/10.48550/arxiv.2204.12632},
\textbf{\texttt{Figure-of-Speech}},\footnote{\url{https://www.kaggle.com/datasets/varchitalalwani/figure-of-speech}}  \textbf{\texttt{CrossLing Metaphors}} \cite{tsvetkov-etal-2014-metaphor}  and  \textbf{\texttt{Metaphor Paraphrase}} \cite{bizzoni-lappin-2018-predicting}.

\begin{table}[!ht]
\small{
\begin{tabular}{|p{7.15cm}|}
\hline
\begin{tabular}[c]{@{}l@{}}{\color{blue}\textit{Your task will be to elaborate a metaphor with rich visual}}\\{\color{blue}\textit{details along with the provided objects to be included}}\\ {\color{blue}\textit{and implicit meaning. Make sure to include the implicit}}\\{\color{blue}\textit{meaning and the objects to be}} {\color{blue}\textit{included in the explanation}} \\ 1. \textbf{Metaphor}: My lawyer is a shark. \\ {\bf Objects to be included}: Lawyer, Shark\\ \textbf{Implicit Meaning}: fierce\\ \textbf{Visual elaboration}: A shark in a suit with fierce\\ eyes \& a suitcase \& a mouth open with pointy teeth.\end{tabular} \\ \hline
\begin{tabular}[c]{@{}l@{}}2. \textbf{Metaphor}: I've reached my boiling point.\\ \textbf{Objects to be included}: Person, Boiling Pot\\ \textbf{Implicit Meaning}: anger\\ \textbf{Visual elaboration}: A boiling pot of water with a\\ person's head popping out of the top, steam coming out \\of their ears, and an angry expression on their face.\end{tabular}                                                                                                                                                                           \\\hline
\begin{tabular}[c]{@{}l@{}}3. \textbf{Metaphor}: Joe: that's because you're like a snail \\surfing on molasses. \\ \textbf{Objects to be included}: Person like a snail, Snail on\\ molasses\\ \textbf{Implicit Meaning}: slow\\ \textbf{Visual elaboration}: A person with a snail shell on their\\ back slowly sliding down a hill of molasses.\end{tabular}                                                                                                                                                                                     \\ \hline
\begin{tabular}[c]{@{}l@{}}4. \textbf{Metaphor}: Absence is the dark room in which lovers \\develop negatives\\ \textbf{Objects to be included}: Darkroom, Negative Film Strip \\with a red heart, Person \\ \textbf{Implicit Meaning}: ominous and lonely\\ \textbf{Visual elaboration}: An ominous dark room with a film \\strip negatives hanging and a red heart in the center with \\a person in the corner looking sad and lonely\end{tabular}                                                                                                         \\ \hline
\begin{tabular}[c]{@{}l@{}}5. \textbf{Metaphor}: My heart is a rose thorn\\  \textbf{Objects to be included}: Heart, Thorn\\ \textbf{Implicit Meaning}: prickly\\ \textbf{Visual elaboration}: A heart with a prickly thorn \\coming out of the center and barbs going outwards.\end{tabular}                                                                                                                                                                                                                                                  \\ \hline\hline
\begin{tabular}[c]{@{}l@{}}6. \textbf{Metaphor}: My bedroom is a pig sty\\  {\color{brown}\textbf{Objects to be included}: Messy bedroom, Pig}\\ {\color{brown}\textbf{Implicit Meaning}: dirty}\\ {\color{brown}\textbf{Visual elaboration}: A bedroom with clothes \& garbage} \\{\color{brown}everywhere with a pig in the center} {\color{brown}rooting around.}\end{tabular} \\   \hline
\end{tabular}
}
\caption{\label{table:prompt} Chain-of-Thought (CoT) prompt to elicit a visual elaboration for a given metaphor. We provide the first five examples in a few-shot learning setting and the model
 jointly generates Objects to be Included, Implicit Meaning, and Visual elaboration (highlighted in brown) step-by-step.}
\end{table}

\paragraph{1) Visually Grounded Linguistic Metaphors:}
Given that not all linguistic metaphors can be rendered as visual metaphors, we manually select those that are visually grounded. Concrete subjects can clearly be visually grounded, but some abstract subjects can be visually grounded as well through their usual representations in media. For example, ``\textit{love}'' can be represented as two people holding hands with hearts above them, ``\textit{confusion}'' as question marks, or ``\textit{idea}'' as a lightbulb over someone's head. Linguistic metaphors that describe non-visual phenomena (e.g., a smell, a sound) are removed unless the act of experiencing the sense is the subject of the sentence, which can be visualized with, e.g., a facial expression. We consider emotional phenomena as visual since often emotions and feelings are expressed through facial expression and/or body posture which can be visualized.

\paragraph{2) Visual Elaboration Generation with Chain-of-Thought Prompting:} Existing text-to-image generation models do not perform well when their input contains linguistic metaphors, since they lack the ability to model implicit meaning and compositionality. Recently, \citet{NEURIPS2022_9d560961} proposed a prompting method for improving the reasoning abilities of language models. This method, called \textbf{Chain-of-Thought} (CoT) prompting, enables models to decompose multi-step problems into intermediate steps. We take advantage of CoT prompting by using the relevant objects and implicit meaning of the metaphors as our intermediate steps, to then elicit detailed textual visualizations of linguistic metaphors using Instruct GPT-3 (davinci-002). We refer to this detailed textual visualization as a \textbf{visual elaboration}. We hypothesize that these visual elaborations obtained from CoT prompting will help text-to-image models create better visual metaphors, as the objects and implicit meaning will  be  explicitly contained in the input.

\begin{figure}[!ht]
    \small
    \centering
    \subfloat[]{\includegraphics[width=0.23\textwidth]{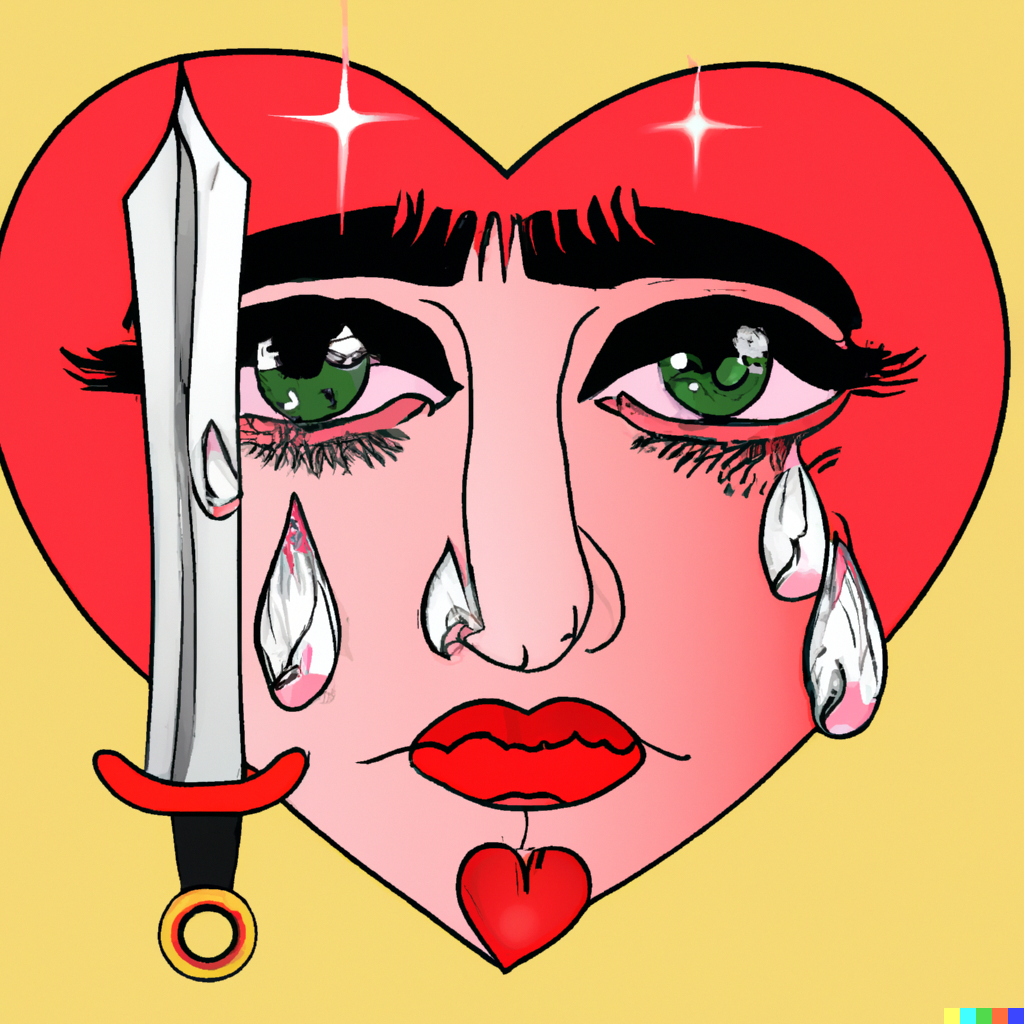}}
    \subfloat[]{\includegraphics[width=0.23\textwidth]{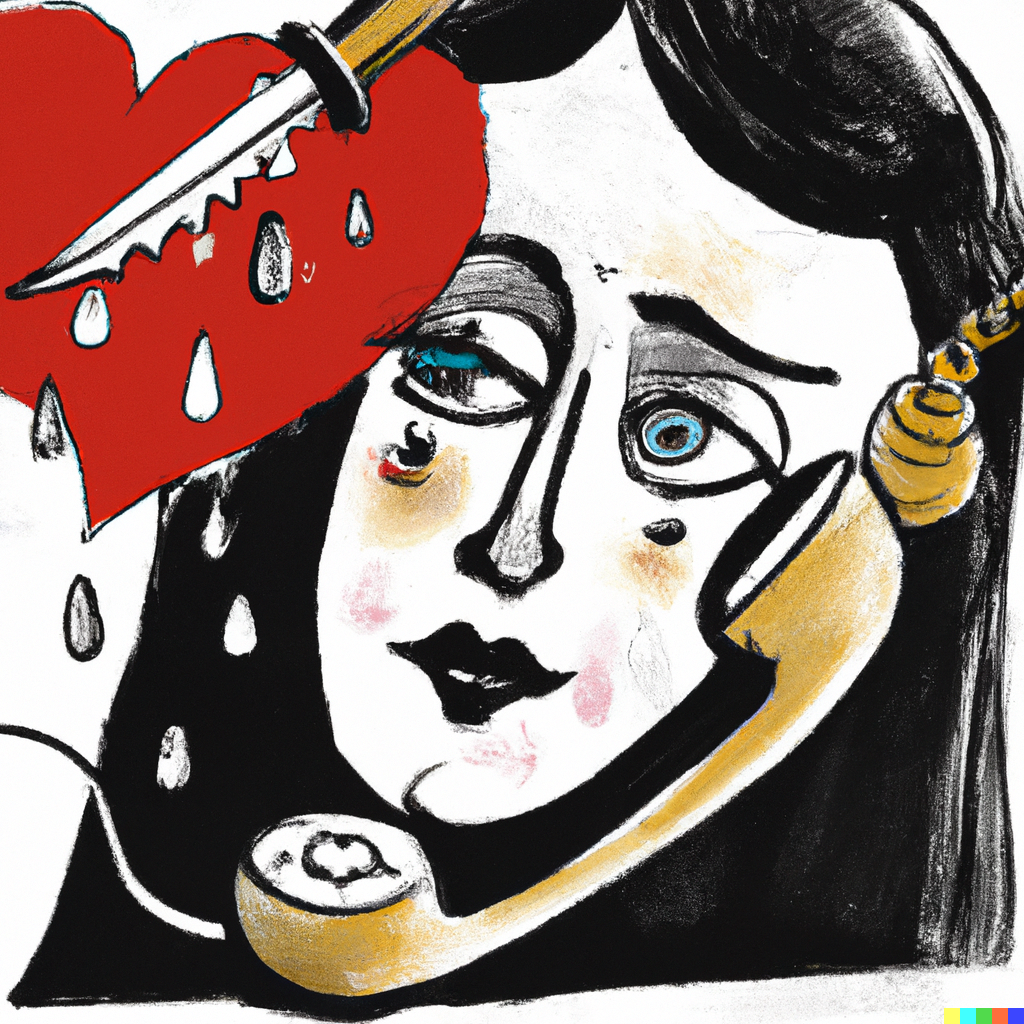}}
    \caption{Visual metaphors obtained using DALL$\cdot$E 2 for the linguistic metaphor ``\textit{The news of the accident was a dagger in her heart}''. The image on the left is obtained using the visual elaboration
    ``\textit{An illustration of a heart with a dagger stuck into it, dripping with blood and pain in the woman's eyes''},
    and the one on the right using the edited prompt ``\textit{An illustration of a {\color{blue}woman receiving a phone call and her} heart with a dagger stuck into it, dripping with blood and pain in the woman's eyes.}''}
    \label{fig:prompt edit}
\end{figure}

Table \ref{table:prompt} shows the instruction and CoT prompt used to elicit a visual elaboration for a given linguistic metaphor. The first five examples are given as few-shot examples and the model (Instruct GPT-3 (davinci-002)) then jointly generates the objects to be Included, implicit meaning, and visual elaboration (highlighted in brown) step-by-step. As our prompts follow a certain structure for step-by-step reasoning, a zero-shot approach would not work well. We found that using five few-shot examples was sufficient to generate elaborations of good quality. We selected five representative examples of visualizable metaphors for the prompt. We used the same examples for generation of every visual elaboration.

While this approach leads to good-quality outputs, not all generated visual elaborations are perfect. We recruit three expert annotators with multiple years of experience in figurative language research and ask them to validate the generated visual elaborations and to slightly edit them if needed, in order to make sure they accurately represent the implicit meaning and the objects involved. Our pipeline is illustrated in Figure \ref{fig:pipeline}. As can be seen in Figure \ref{fig:prompt edit}, for the given linguistic metaphor ``\textit{The news of the accident was a dagger in her heart}'', the first visual elaboration is almost correct but it misses the crucial information about the metaphoric source, i.e ``\textit{the news of the accident}''. An expert performs a minor edit by adding the phrase ``\textit{woman receiving a phone call}'' in order to convey the metaphoric source which leads to a perfect visual metaphor. Experts performed minor edits on 29\% of the generated visual elaborations.

\paragraph{3) Visual Metaphor Generation using Diffusion-based Models and Human Quality Check:}
For this part of the data curation process, we first prompt DALL$\cdot$E 2 to generate multiple images\footnote{DALL$\cdot$E 2 automatically generates four images per prompt.} for a single visual elaboration (cf.~ Figure \ref{fig:pipeline}). Post-generation, each set of generated images is examined jointly by three experts to determine whether they accurately and fully represent the meaning of the original linguistic metaphor. The experts need to validate whether the image contains the relevant objects and whether the objects are positioned correctly or have the appropriate indicators of movement or action, also referred to as \textbf{Attribute Binding} \cite{ramesh2022hierarchical,saharia2022photorealistic}. For example, for the phrase ``\textit{Her eyes were like peonies}'', the image would need to depict both a face and peonies and the peonies would need to be in the place of the eyes rather than around the head (which was the case in some images). Images that do not meet the above criterion were discarded. 

The dataset curated in this way contains \textbf{1,540} unique linguistic metaphors (and their associated visual elaborations) and \textbf{6,476} unique images. Each linguistic metaphor
has \textbf{four} associated images, on average. 
We call our data \textbf{\texttt{HAIVMet}} (\textbf{H}uman-\textbf{AI} \textbf{V}isual \textbf{Met}aphor). 

\section{Evaluation}
\label{sec:eval}
Our goal is to assess the impact of the LLM-Diffusion Model collaboration (Section \ref{models}), and of the Human-AI collaboration on building a high-quality dataset.

\subsection{LLM-Diffusion Model Collaboration} \label{models}

\paragraph{Models.}
Diffusion models are trained to recover the original version of an image after random noise has been applied to it \citep{ramesh2022hierarchical}. Both DALL$\cdot$E 2 and Stable Diffusion are diffusion-based text-to-image models. Stable Diffusion is open source; DALL$\cdot$E 2 is not. Note that in this evaluation, there is no human intervention (no editing of the output of Instruct GPT-3 with CoT prompting, nor filtering of images produced by diffusion-based models). We use the following LLM-Diffusion Model collaboration setups, where the input to the diffusion models is the visual elaboration of the linguistic metaphor generated using Instruct GPT-3 (davinci-002) with CoT prompting:

\begin{itemize}[leftmargin=*]
    \itemsep0em 
    \item \textbf{LLM-DALL$\cdot$E 2}: DALL$\cdot$E 2  \citep{ramesh2022hierarchical} with the LLM-generated visual elaboration as input. 
    \item \textbf{LLM-SD}: The Stable Diffusion \citep{Rombach_2022_CVPR} v2.1 model, with the same input as LLM-DALL$\cdot$E 2.
    \item \textbf{LLM-SD$_{Structured}$} We use the diffusion method of \citet{feng2022training} which combines the structured representations of prompts (for example, their constituency tree) with the diffusion guidance process, using the same input as LLM-DALL$\cdot$E 2.
\end{itemize}

We also use DALL$\cdot$E 2 and Stable Diffusion (\textbf{SD}) with the linguistic metaphor given directly as input (no collaboration with the LLM). This comparison allows us to assess the benefit that can be drawn from LLM-Diffusion Model collaboration.

\begin{figure*}
    \includegraphics[width=\textwidth]{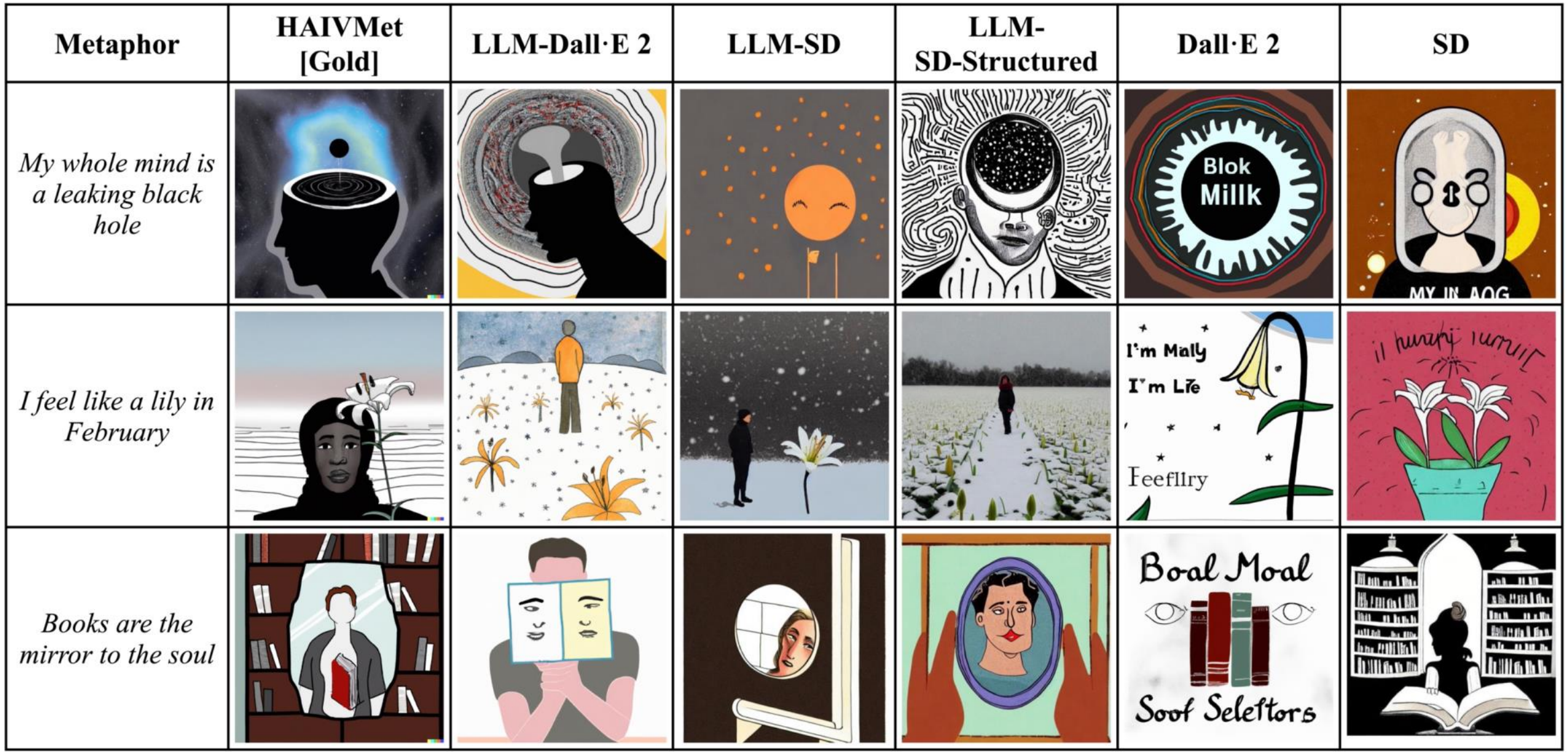}
    \caption{Examples of output from each model 
    described in Section \ref{models} for three randomly chosen metaphors. \textbf{\texttt{HAIVMet}} is our gold standard. More examples are available in Figure \ref{fig:more_qual} in the Appendix.\label{fig:goldtable}}
\end{figure*}

\paragraph{Human Evaluation Setup.}
\label{sec:human_eval}
Among the popular automatic evaluation metrics, both Fréchet Inception Distance (FID) \cite{heusel2017gans} and CLIP \cite{pmlr-v139-radford21a} scores are not tailored towards metaphorical images, and are not reliable in assessing whether the generated images capture the essence of visual metaphors \cite{https://doi.org/10.48550/arxiv.2212.09898}. We also chose not to rely on non-expert crowd-workers as even with training they have been found to be unreliable for open-ended tasks \cite{karpinska2021perils}. Following the recommendation from \citet{karpinska2021perils}, we recruit three professional artists with experience in concept illustration and visual arts through the Upwork\footnote{\url{https://www.upwork.com}} platform.
We ask them to evaluate the visual metaphors that are generated by the five approaches described above for a subset of 100 randomly selected linguistic metaphors from our dataset. For each metaphor, we ask to rank the five generated images on the basis of how well they represent the metaphor.

Additionally, we collect targeted feedback by asking the raters to provide natural language instructions for improving the images. Five text fields are shown under each image, and  the annotators are invited to make up to five recommendations. In the occasional case where the image is ``Perfect'' or absolutely not worthy of transformation (``Lost Cause''), the annotators do not need to provide any feedback for improvement. 
\begin{figure*}[!ht]
    \includegraphics[width=\textwidth]{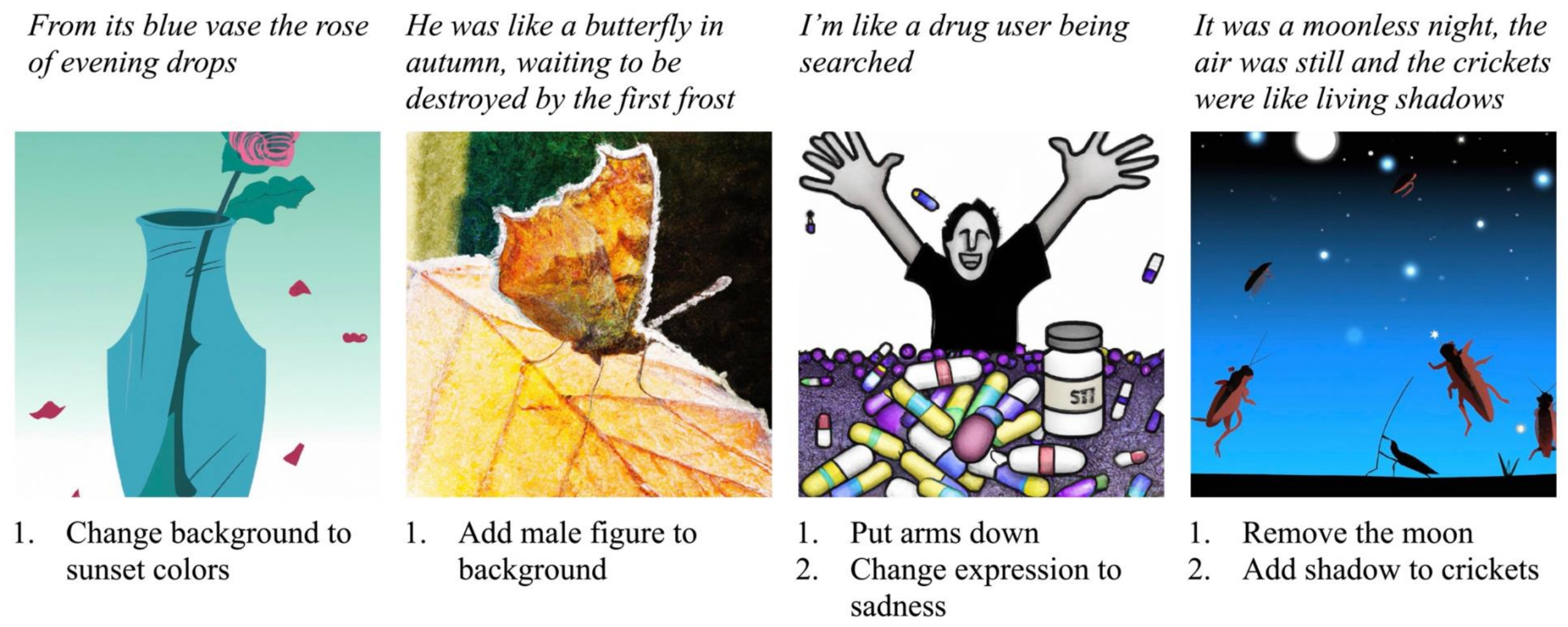}
    \caption{\label{edits} 
    Experts suggested image edits in the form of natural language instructions for images  generated from the CoT visual elaborations of linguistic metaphors.}
    \vspace{-2ex}
\end{figure*}
 The suggested types of instructions are the following: 1) Add an object; 2) Remove an object; 3) Move an object; 4) Replace an object with another object; 5) Change an object’s property (e.g., color, size). The annotators are encouraged to supply whatever type of change they believe is required to improve the visual metaphor; the only stipulation to the instructions is that each one must denote a single action/change. We identify the average rank assigned to a model across metaphors and annotators. We also report the percentage of ``Lost Cause'' cases in order to identify systems that generate the least amount of bad images. Additionally, we compare the models on the basis of the average number of instructions that have been proposed for improving their produced images. The number of suggested changes acts as a proxy for how close the image is to the perfect representation of the metaphor. ``Perfect'' images are considered to have 0 edits, and images that are a ``Lost Cause'' are considered to have 5 edits to ensure fairness in this computation.

\begin{table}[ht]
\small
\centering
\begin{tabular}{|l|l|l|l|}
\hline
\textbf{Model} & \textbf{\begin{tabular}[c]{@{}l@{}}Avg \\ Rank\end{tabular}} & \textbf{\begin{tabular}[c]{@{}l@{}}\% Lost \\ Cause\end{tabular}} & \textbf{\begin{tabular}[c]{@{}l@{}}Avg \# of \\ Instructions\end{tabular}} \\ \hline
SD & 3.82 & 31.6 & 2.25  \\ \hline
LLM-SD & 3.40 & 23.3 & 1.83  \\ \hline
LLM-SD$_{Structured}$ & 3.05  & 18.3  & 1.57 \\ \hline
DALL$\cdot$E 2 & 2.76 & 16.6 & 1.44  \\ \hline 
\begin{tabular}[c]{@{}l@{}}LLM-DALL$\cdot$E 2\end{tabular} & \textbf{1.96} & \textbf{6.0} & \textbf{0.76} \\ \hline
\end{tabular}
\caption{\label{table:human_res} Human evaluation results: 
the average ranking given by three human raters to the output of each model for 100 test metaphors; the percentage of images labeled as ``Lost Cause''; and the average number of edits needed to make the image perfect otherwise.}
\end{table}

\subsubsection{Results and Analysis}
Table \ref{table:human_res} shows that without collaboration with a LLM (i.e., just with the linguistic metaphor as input), DALL$\cdot$E 2 performs better than SD (line 4 vs. line 1). The main take away is that LLM-Diffusion Model collaboration outperforms simple Diffusion Models (LLM-DALL$\cdot$E 2 vs. DALL$\cdot$E 2, LLM-SD and LLM-SD$_{Structured}$ vs. SD). That is, using Instruct-GPT3 with CoT prompting to produce visual elaborations as input to diffusion models consistently improves the performance over providing the diffusion models directly with linguistic metaphors. Overall, LLM-DALL$\cdot$E 2 emerges as the best system. Only 6\% are ``Lost Cause'' images, affirming our choice for using LLM-DALL$\cdot$E 2 to create \texttt{\textbf{HAIVMet}}. Rank 1 (best) was assigned to LLM-DALL$\cdot$E 2 in 44.6\% of cases, followed by 24.0\% for DALL$\cdot$E 2, 14.0\% for LLM-SD$_{Structured}$, 10.0\% for LLM-SD, and 7.3\% for SD. Using the same prompts as for LLM-DALL$\cdot$E 2, we still observe an improvement in LLM-SD over the original SD output. Finally, as expected, LLM-SD$_{Structured}$ improves over LLM-SD. 

In Figure \ref{fig:goldtable}, we show examples of visual metaphors generated using  the linguistic metaphors  or their visual elaborations as  CoT prompts. We observe that the latter, where CoT prompting is involved, are of higher quality. For instance, a good visual metaphor for the metaphorical expression ``\textit{Books are the mirror to the soul}''  would require books, a mirror, and superimposing the mirror with some approximate depiction of a soul (usually illustrated as a person). However, the images that DALL$\cdot$E 2 and Stable Diffusion generate (columns 3 and 5, respectively), just contain books. This problem is fixed with CoT prompting, as seen in columns 2, 4, and 6. The observations are similar for 
the metaphor ``\textit{I feel like a lily in February}'', where the implicit meaning of being \textit{out of place} is depicted by lilies blooming in February over a snowy (instead of sunny) landscape.

\paragraph{How do expert illustrators perceive model-generated visual metaphors?}

One of the goals of our evaluation, besides obtaining a subjective ranking of the tested models, was to analyze some of the flaws in the output. As stated above, for every image that was not considered ``Perfect'' or ``Lost Cause'', we collected suggestions from experts about changes that would improve the image as a visual metaphor. Examples are given in Figure \ref{edits}. This helps us understand where models might still be lacking, and the extent to which future interaction with illustrators might shape model-generated outputs to be acceptable. We find that issues in the output may be due to a model not being able to accurately depict a prompt, due to under-specification in terms of the objects to be represented or to the implicit property not being properly depicted. For instance, the CoT prompt for the metaphor ``\textit{It was a moonless night, the air was still and the crickets were like living shadows}'' accurately describes it as ``\textit{An illustration of a moonless night sky with still air and crickets crawling around as living shadows.}''. However, the model fails to understand the word moonless and adds a moon to the picture. Additionally, while it adds the crawling crickets to the picture, there are no shadows. This affects the way  we perceive the metaphor since its implicit meaning  is ``dark and creepy''. However, the rest of the image is high quality in terms of depiction. On the contrary, for the metaphor ``\textit{He was like a butterfly in autumn, waiting to be destroyed by the first frost}'', the CoT prompt ``\textit{An illustration of a butterfly perched on an autumn leaf with the first frost starting to form around it}'' misses out on the source `He' (ideally a fragile man) but the model depicts it perfectly.

Table \ref{table:human_res} shows that nearly all models have room for improvement. Future work can use these suggestions in the form of natural language instructions to edit model-generated images, as demonstrated in recent work by \citet{brooks2022instructpix2pix}.

\subsection{Human-AI Collaboration Evaluation} \label{haieval}

\paragraph{Intrinsic Evaluation.} 
To better understand if Human-AI collaboration leads to better quality visual metaphors, we conduct another round of evaluation with the same group of professional artists. Our experimental setup is the same as in  our previous evaluation, except that instead of five images, we provide them with two visual metaphors for the same input: one from the \texttt{\textbf{HAIVMet}} corpus and the other from LLM-DALL$\cdot$E 2 used in the previous round of evaluation (with their order shuffled). We then ask them to objectively provide a ranking between the two systems or tie them if they are both of the same quality. They are also asked to provide instructions for improving them (unless they are Perfect or Lost Cause). We get the final verdict using majority voting. We obtain an inter-annotator agreement of 0.57 based on Fleiss’s kappa~\cite{fleiss1971measuring} (``moderate agreement''). Our results in 
Table \ref{table:human_res1} show that while 37\% of the images are of similar quality, from the remaining images professionals preferred instances from \texttt{\textbf{HAIVMet}} 45\% of the time compared to LLM-DALL$\cdot$E 2 18\% of time. Finally, the \texttt{\textbf{HAIVMet}} data has an almost negligible number of Lost Causes, providing further evidence of its high quality. 
\begin{table}[ht]
\small
\centering
\begin{tabular}{|l|l|l|l|}
\hline
Criterion & LLM-DALL$\cdot$E 2 & \texttt{\textbf{HAIVMet}} & Tie \\ \hline
Preference & 18\%         & \textbf{45.0}\%     & 37\%  \\ \hline
Lost Cause & 5\%         & \textbf{1.6}\%     & -  \\ \hline
Perfect & 52\%         & \textbf{63.6}\%     &  -\\ \hline
\end{tabular}
\caption{\label{table:human_res1} Proportion of Preference, Lost Cause, and Perfect
cases from LLM-DALL$\cdot$E 2 and \texttt{\textbf{HAIVMet}} for  metaphors in our blind test set.}
\end{table}

\paragraph{Extrinsic Evaluation: Visual Entailment Task.}
\begin{figure}
    \centering
\includegraphics[width=\linewidth]{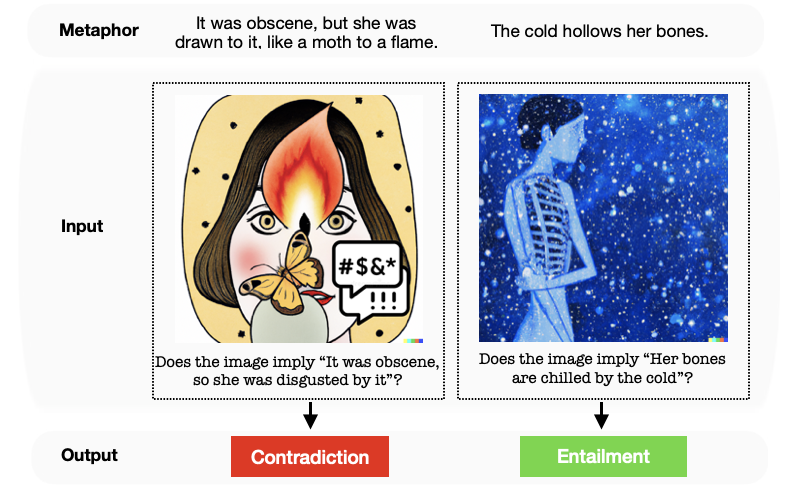}
    \caption{Visual Entailment task. Given an image and a prompt containing a hypothesis, predict whether the hypothesis entails, contradicts, or is neutral to the image.}
    \label{fig:ve_example}
\end{figure}

\begin{table}[!t]
\centering
\begin{tabular}{|ccc|}
\hline
\multicolumn{1}{|c|}{Model}
&Dev & Test \\ \hline
\multicolumn{1}{|c|}{OFA$_{\text{SNLI-VE}}$}   &    25.25  &   27.81  \\
\multicolumn{1}{|c|}{OFA$_{\text{SNLI-VE+\texttt{\textbf{HAIVMet}}}}$}   &  \textbf{49.90}  & \textbf{51.15}\\\hline

\end{tabular}
 \vspace{-1.5ex}
\label{tab:ve_results}
\caption{Visual Entailment Results. OFA \cite{wang2022ofa} fined-tuned on SNLI-VE \cite{xie2019visual} vs. SNLI-VE+\texttt{\textbf{HAIVMet}}. Bold indicates best performance.} 
\end{table}

Apart from being a rich source of visual metaphors, our dataset can also be useful in downstream applications. We showcase this by using it in a Visual Entailment (VE) task, where a vision-language model needs to predict whether a hypothesis is entailed by an image  (cf. Figure \ref{fig:ve_example}). We use OFA \cite{wang2022ofa}, a state-of-the-art VE model finetuned on SNLI-VE \cite{xie2019visual}. SNLI-VE  only contains real-world images, but OFA is pre-trained on $\sim$20M image-text pairs some of which are synthetic.  We extract 958 metaphors from our dataset that are associated with \textbf{literal} natural language entailment pairs from FLUTE~\cite{chakrabarty2022flute}, CrossLing Metaphors~\cite{tsvetkov-etal-2014-metaphor} and Metaphor Paraphrase~\cite{bizzoni-lappin-2018-predicting} (see Appendix \ref{app:viz_entailment} for details on the data construction procedure). We split the data into train, validation and test sets, which contain 708, 100 and 150 metaphors (3686/506/831 image-text pairs), respectively. We fine-tune OFA-base (182M parameters) for 10 epochs with learning rate 6e-5 and polynomial decay (weight=0.01), and batch size 8 on an NVIDIA RTX A6000 GPU for 8 hours. We select the model that has best performance on the development set. We show that accuracy on the test set improves by $\sim$23 points compared to OFA's performance when it is only finetuned on SNLI-VE. This result is indicative of the quality and usefulness of our dataset which can help vision-language models capture metaphoric meaning

\section{Compositionality in Visual Metaphors}

\begin{table}[!ht]
\small
\centering
\begin{tabular}{m{0.45\linewidth} m{0.45\linewidth}}
\hline
\centering He froze with fear when he saw it. & \centering\arraybackslash \includegraphics[width=1.0\linewidth]{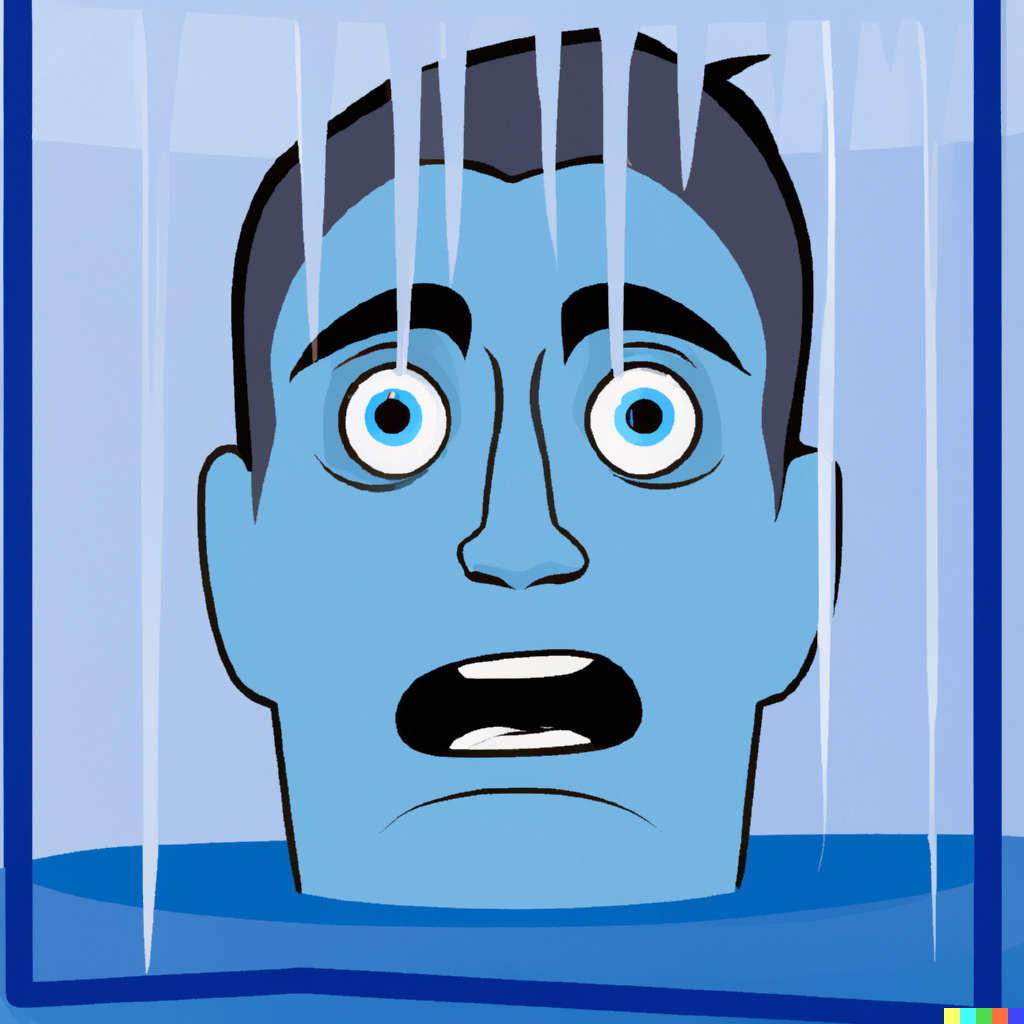} \\
\hline
\centering Beauty is a fading flower. & \centering\arraybackslash \includegraphics[width=1.0\linewidth]{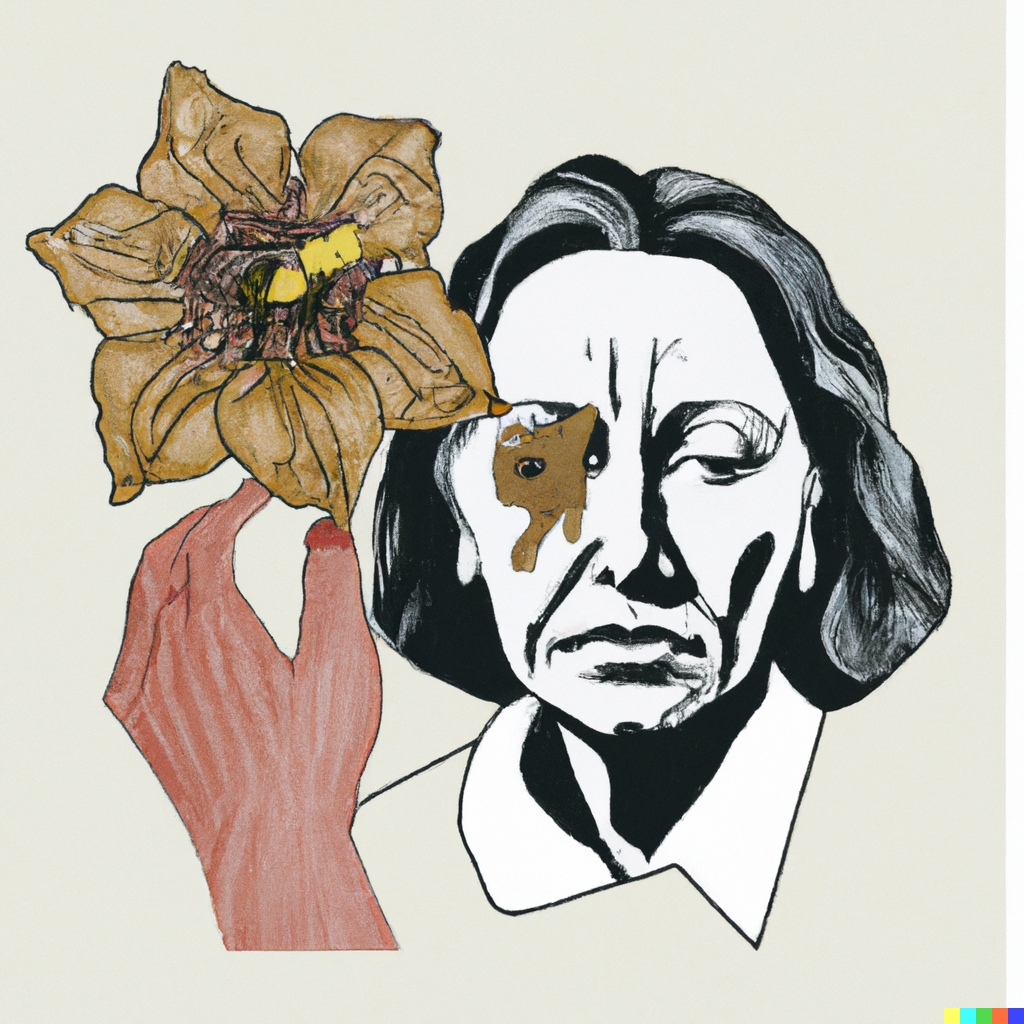} \\
\hline
\centering Love is a crocodile in the river of desire. & \centering\arraybackslash \includegraphics[width=1.0\linewidth]{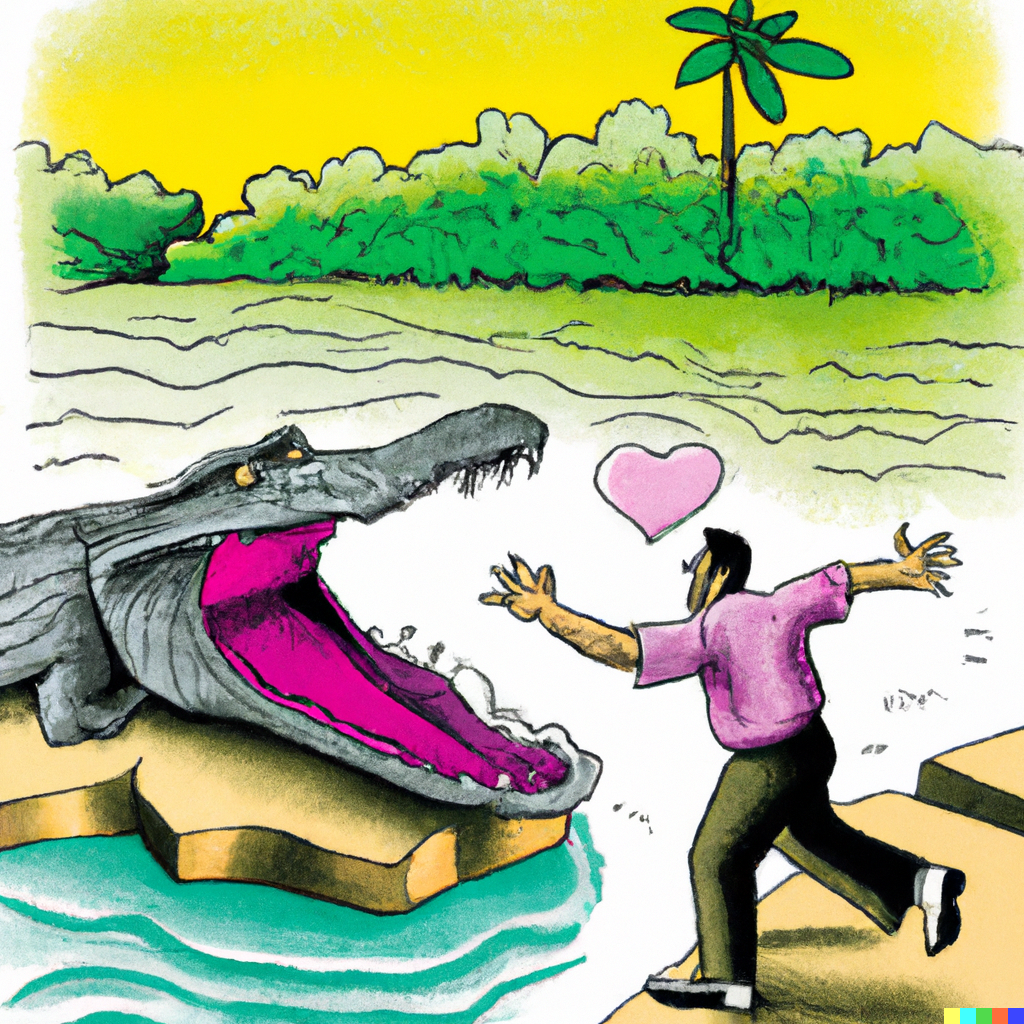} \\
\hline
\end{tabular}
\caption{\label{comp}Visual Metaphors from the \texttt{\textbf{HAIVMet}} dataset demonstrating compositional generalization.}
\label{tab:gt}
\end{table}

In prior work, \citet{gutierrez-etal-2016-literal} showed that metaphorical meaning is not only a property of individual words but arises through cross-domain composition. \citet{Gal2019-GALVMR} further argues that a metaphor is  a visual material rather than conceptual. It is a mechanism of syntactic structure, forms, and material composition, which goes along with the perception of structures and compositions. Many images from our HAIVMet data showcase the compositional nature of visual metaphors, as can be seen in Table \ref{comp}. For example, to visualize the metaphor ``\textit{Love is a crocodile in the river of desire}'' the model needs to show both a human and a crocodile while depicting a sense of desire by embodying love as a concept. Similarly, for ``\textit{He froze with fear when he saw it}'', the metaphor needs to not only depict fear but also combine it with the state of being frozen. We can successfully achieve these difficult compositional visualizations through efficient human-AI collaboration.

\section{Conclusion}

We show that using Chain-of-Thought prompting for generating visual elaborations of linguistic metaphors leads to significant improvements in the quality of visual metaphors generated by diffusion-based text-to-image models. These models excel at depicting literal objects and actions, but cannot make the leap from figurative phrases to visual depiction without a detailed explanation of the implicit meaning. Though there are still particular aspects of visual composition and figurative imagery that current models fail to capture, the breadth of information collected in this dataset not only allows us to understand the current limitations of image generation but also provides the data necessary to improve visual metaphor generation in the future. We plan to further examine the effect of  prompt phrasing on the quality of the generated visual metaphors,and how that effect differs across different models.

\section*{Limitations}

While the results of Human-AI collaboration for visual metaphor generation are very promising, such a procedure might be time-consuming but at the same time necessary for maintaining quality. We want to acknowledge that both our LLM and best-forming Diffusion models are released through a paid API and are not open-sourced. 
While our best-performing system uses Chain Of Thought Prompting,  there are several other prompting or task decomposition techniques that we did not perform an extensive comparison with.Last but not least, there is still enough room for potential improvement in generating visual metaphors which can be achieved by designing better prompts or by improving the compositional generalization of diffusion models. We also recognize the inherent limitation of an English-only basis for our visual metaphors and hope in the future to expand to other languages for source material.

\section*{Ethics Statement}

The use of text-to-image generation models is subject to concerns about intellectual property and copyrights of the images generated since the models are trained on web-crawled images. Our task is restricted to generating visual metaphors from linguistic metaphors, and the human-AI collaboration setup should be considered as a creative aid tool. All data collected by human respondents were anonymized and only pertained to the data they were being shown. We do not report demographic or geographic information, given the limited number of respondents, so as to maintain full anonymity. Workers on UpWork were informed that that the work they were doing was going to be used for research purposes. They were paid a wage of 20\$ per hour as decided by the workers themselves. Workers were paid their wages in full immediately upon the completion of their work.

\bibliography{custom}
\bibliographystyle{acl_natbib}
\clearpage
\appendix
\section{Appendix}
\label{sec:appendix}

\subsection{Hyperparameters for chain of Thought Prompting}

We use the Instruct GPT-3 (davinci-002) model for Chain-of-Thought (CoT) prompting. To generate \textbf{Objects to be Included}, \textbf{Implicit Meaning} and \textbf{Visual Elaboration} we use the following hyperparameters: \texttt{temperature=0.7,max tokens=256,top p=1.0,best of=1,
frequency penalty=0.5,presence penalty=0.5.}

\begin{table*}[ht]
\centering
\small
\begin{tabular}{|l|}
\hline
\begin{tabular}[c]{@{}l@{}} 1. \textbf{Metaphor}: My lawyer is a shark.\\
\textit{An illustration of} a shark in a suit with fierce eyes 
and a suitcase and  a mouth open with pointy teeth  \end{tabular} \\ \hline

\begin{tabular}[c]{@{}l@{}} 2. \textbf{Metaphor}: I've reached my boiling point.\\
\textit{ An illustration of} a boiling pot of water  with a person’s head popping  out of the top, \\ steam coming out of their ears,
and an angry expression on  their face. \end{tabular}                                                                                                                      \\\hline
\begin{tabular}[c]{@{}l@{}} 3. \textbf{Metaphor}: Joe: that’s because you’re like a snail surfing on molasses.\\
\textit{An illustration of} a person with a snail shell on their back slowly   sliding down a hill of molasses. \end{tabular}                                                                             \\ \hline
\begin{tabular}[c]{@{}l@{}} 4. \textbf{Metaphor}: Absence is the dark room in which lovers develop negatives.\\
\textit{An illustration of} an ominous dark room with a film strip negatives  hanging and a red heart \\ in the 
center with a person in the corner looking sad and lonely. \end{tabular}                                                                                                 \\ \hline
\begin{tabular}[c]{@{}l@{}} 5. \textbf{Metaphor}: My heart is a rose thorn. \\ 
\textit{An illustration of} a heart with a prickly thorn coming out of the center and barbs going outward.\end{tabular}                                                     \\ \hline
\hline\hline
\begin{tabular}[c]{@{}l@{}} 6. \textbf{Metaphor}: My bedroom is a pig sty\\  {\color{brown}\textit{An illustration of} a messy bedroom with clothes and garbage strewn about and a pig in the center rooting through the mess.}\end{tabular} \\   \hline
\end{tabular}
\caption{\label{table:prompt_no_cot} Simple Completion prompt to elicit visual elaboration for a given metaphor, using the same 5 few shot examples as in the CoT prompting strategy, but without the objects to be included and the implicit meaning.}
\end{table*}

\begin{table}[!ht]
\begin{tabular}{ll}
\textbf{CoT}                    & \textbf{Completion}                  \\
\multicolumn{2}{l}{I had already planted the idea in her mind.}        \\ \includegraphics[width=.4\linewidth]{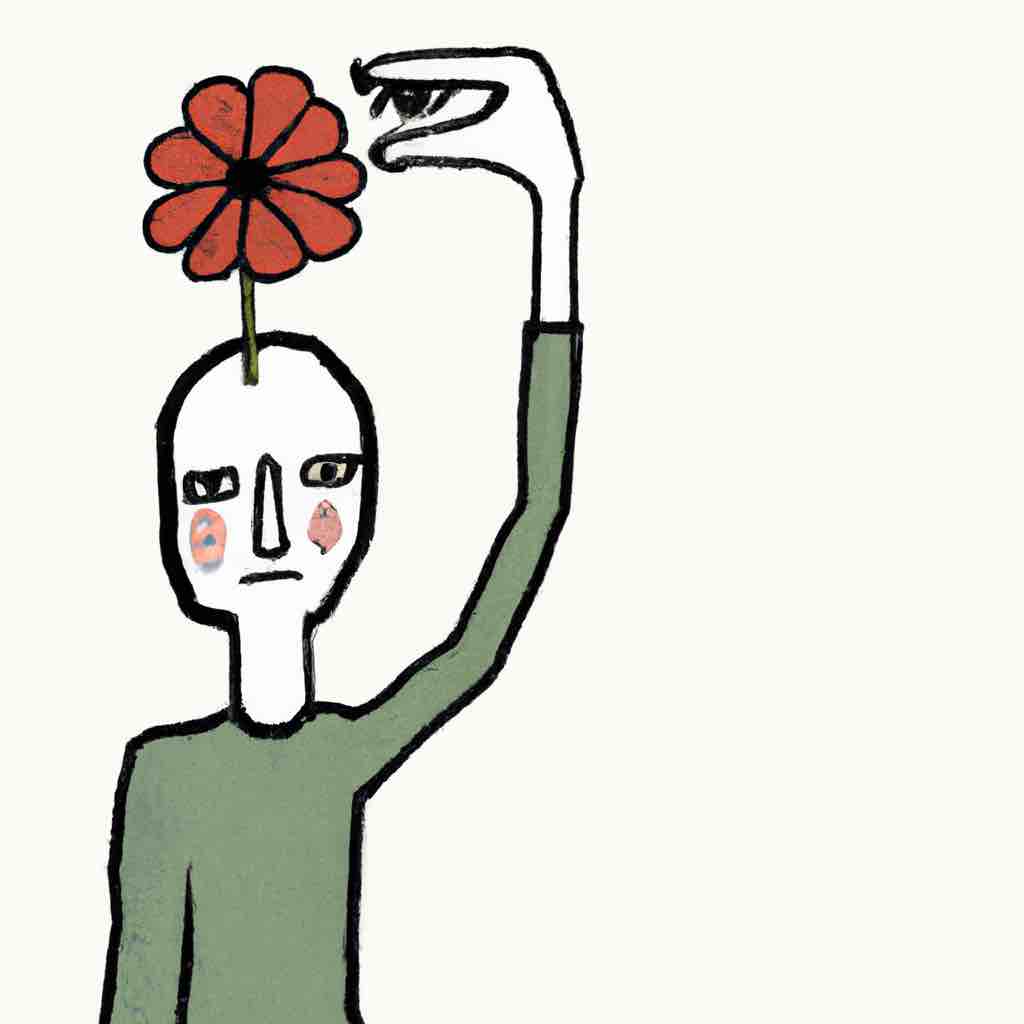}
                                &                 \includegraphics[width=.4\linewidth]{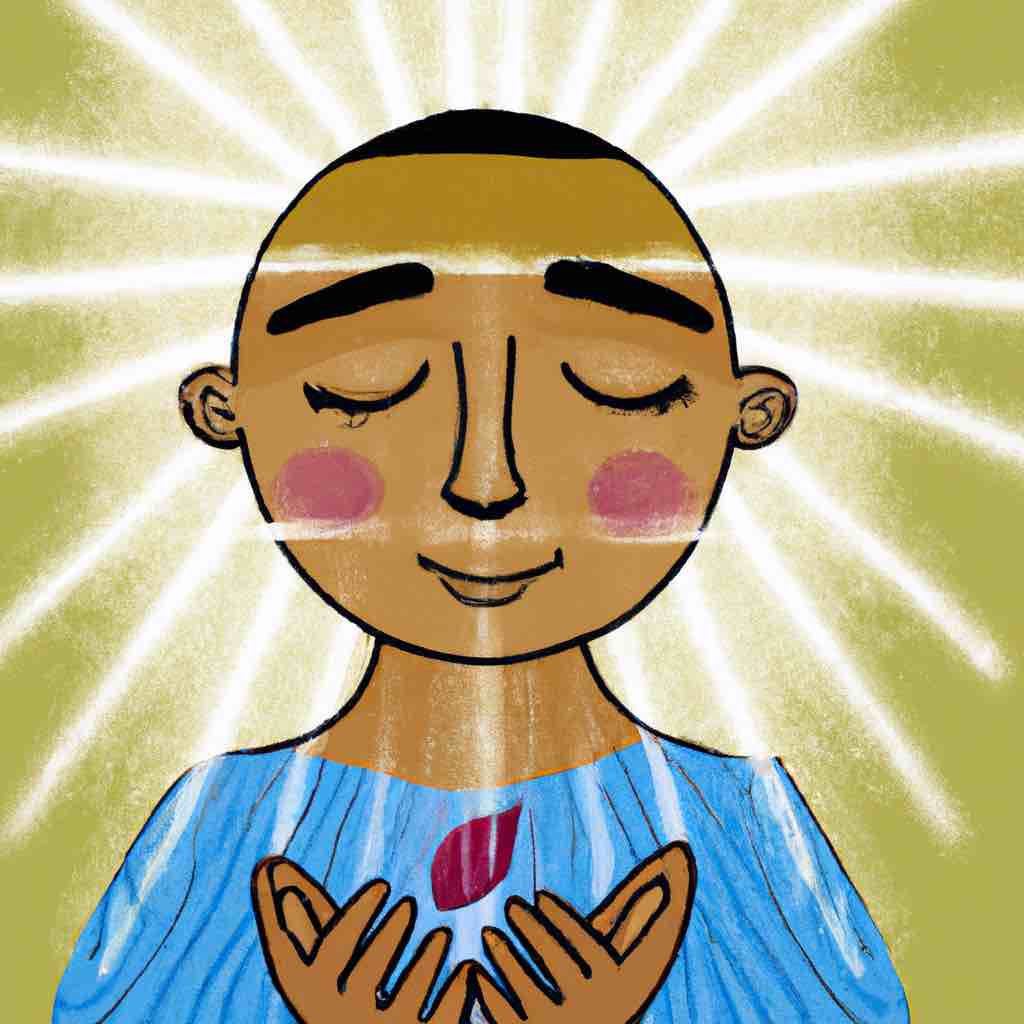}              \\
 \multicolumn{2}{l}{The rooms communicated.}                            \\\includegraphics[width=.4\linewidth]{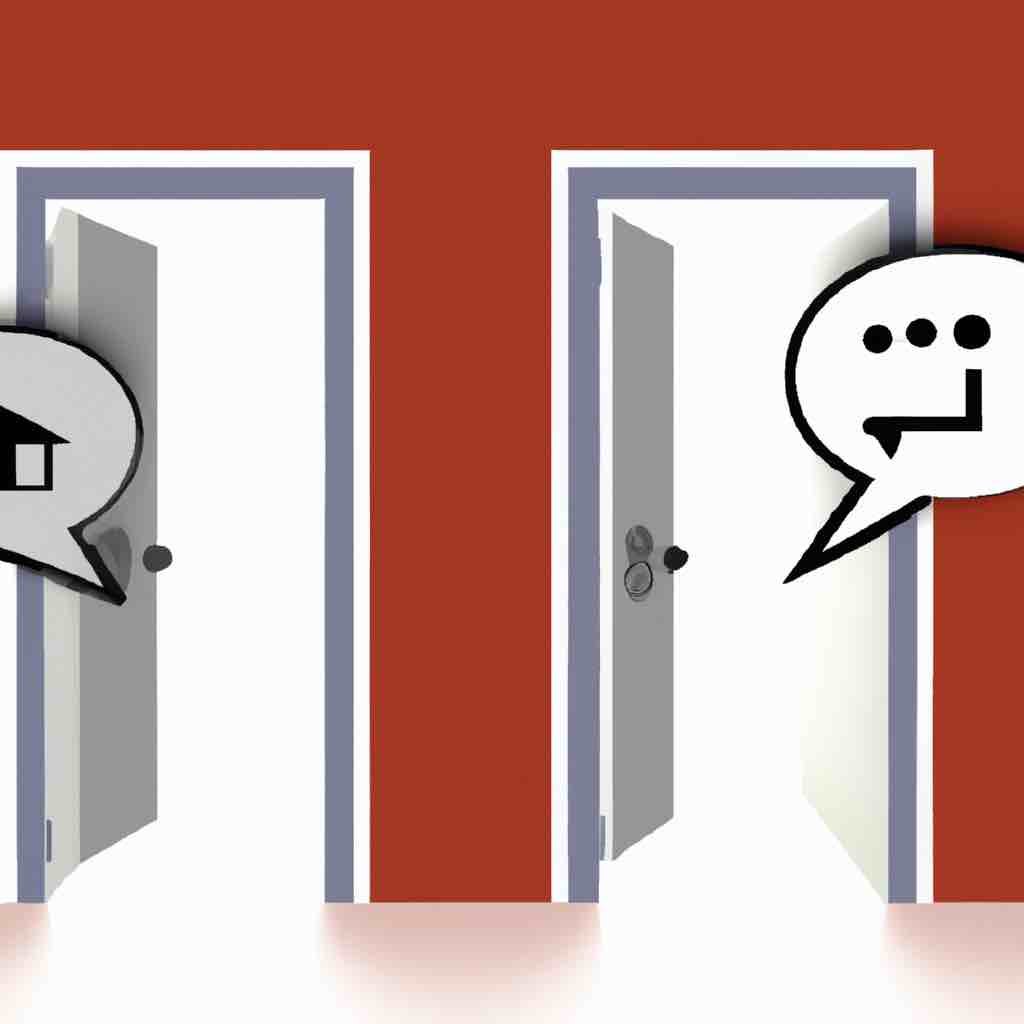}                             &    \includegraphics[width=.4\linewidth]{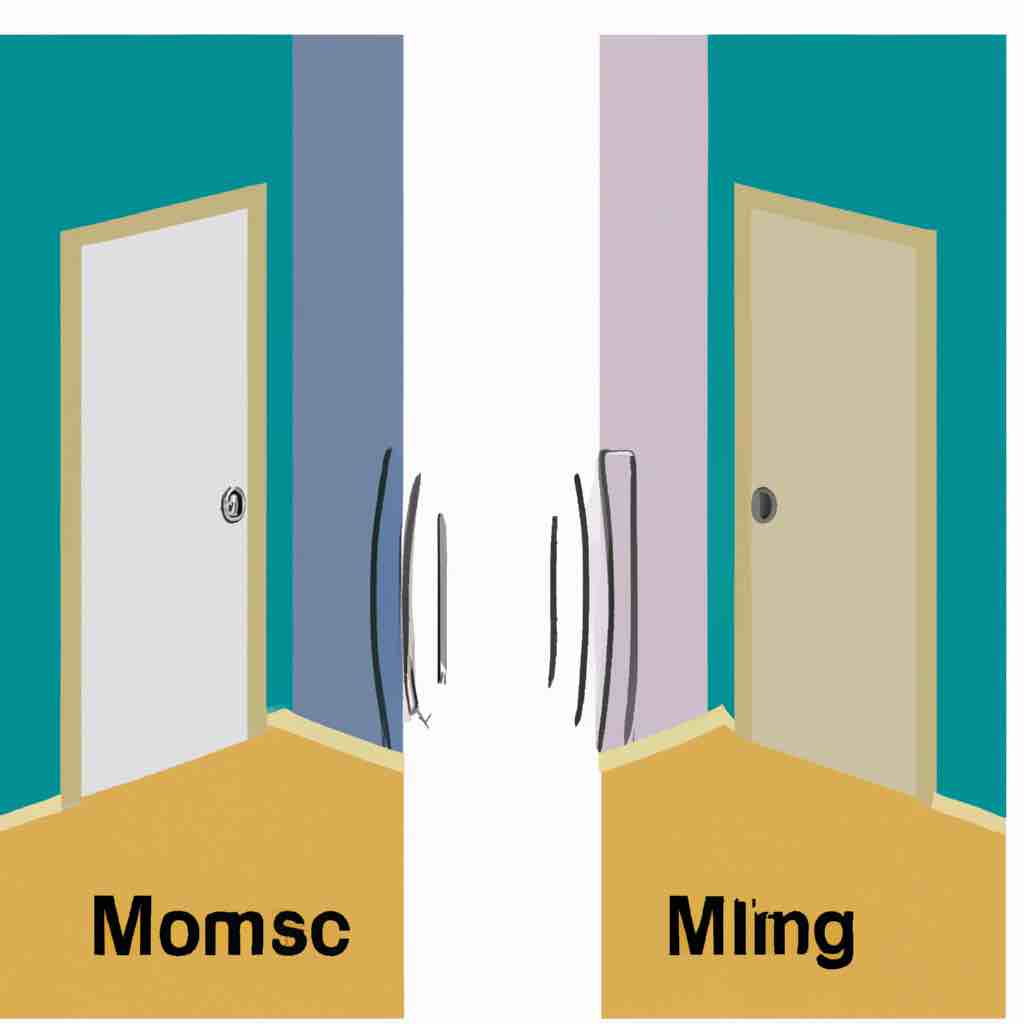}                                  \\
\multicolumn{2}{l}{My big brother is a couch potato.}                  \\
\includegraphics[width=.4\linewidth]{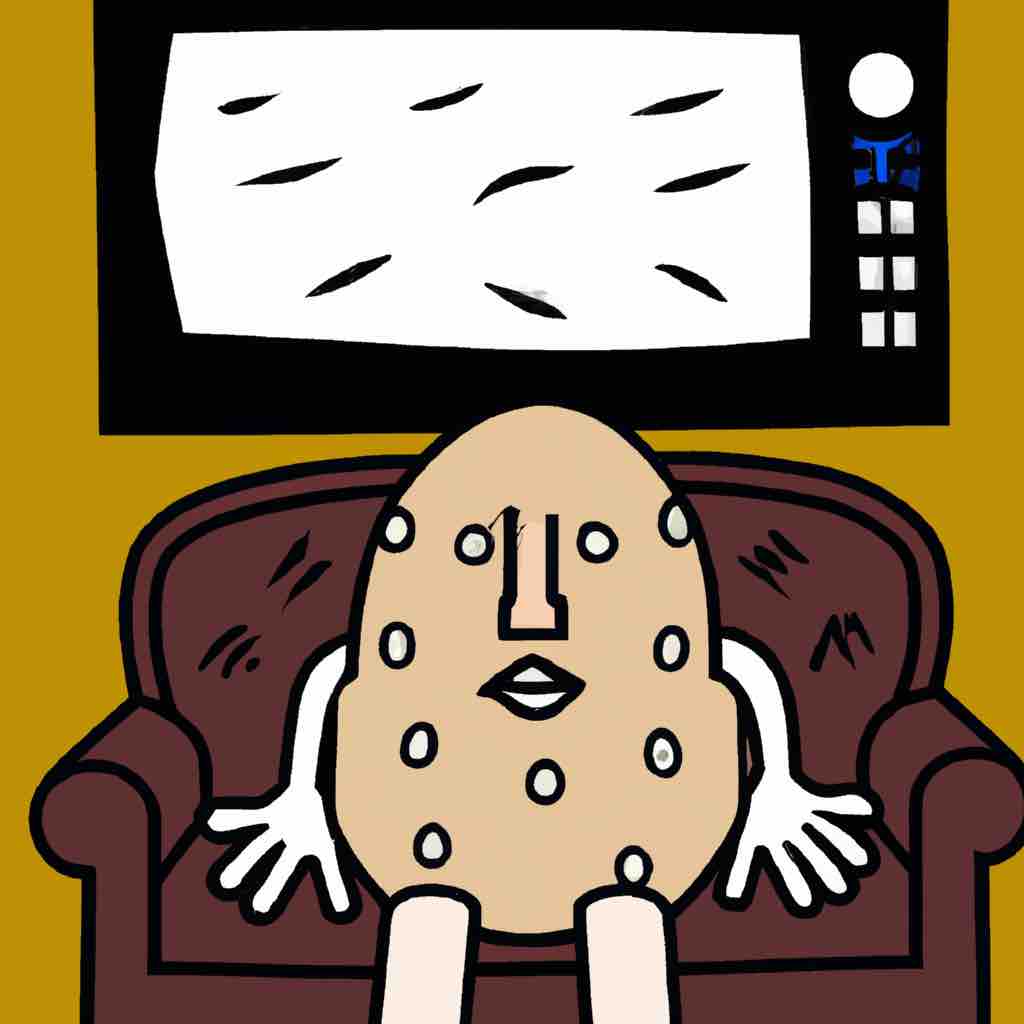}                            &   \includegraphics[width=.4\linewidth]{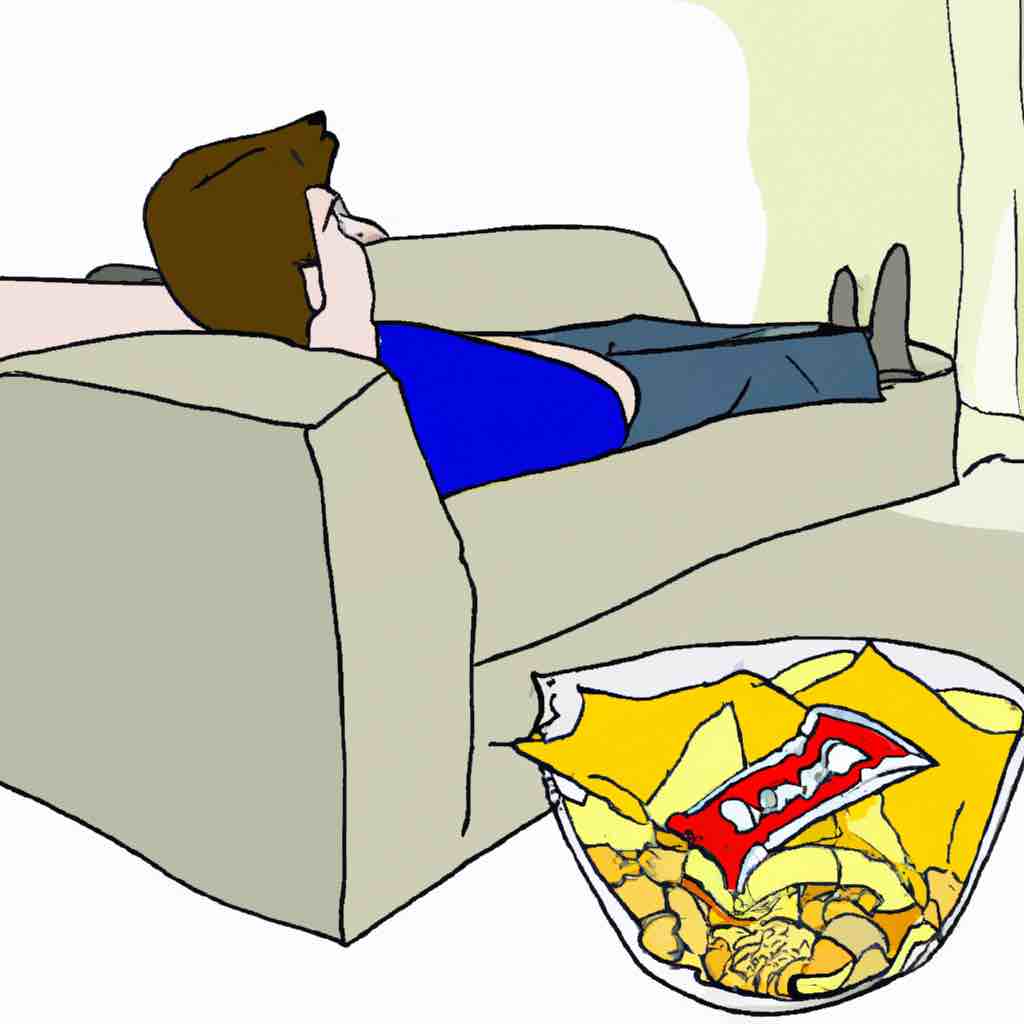}                                    \\
\multicolumn{2}{l}{You will love the new train. It is a heavenly ride} \\
\includegraphics[width=.4\linewidth]{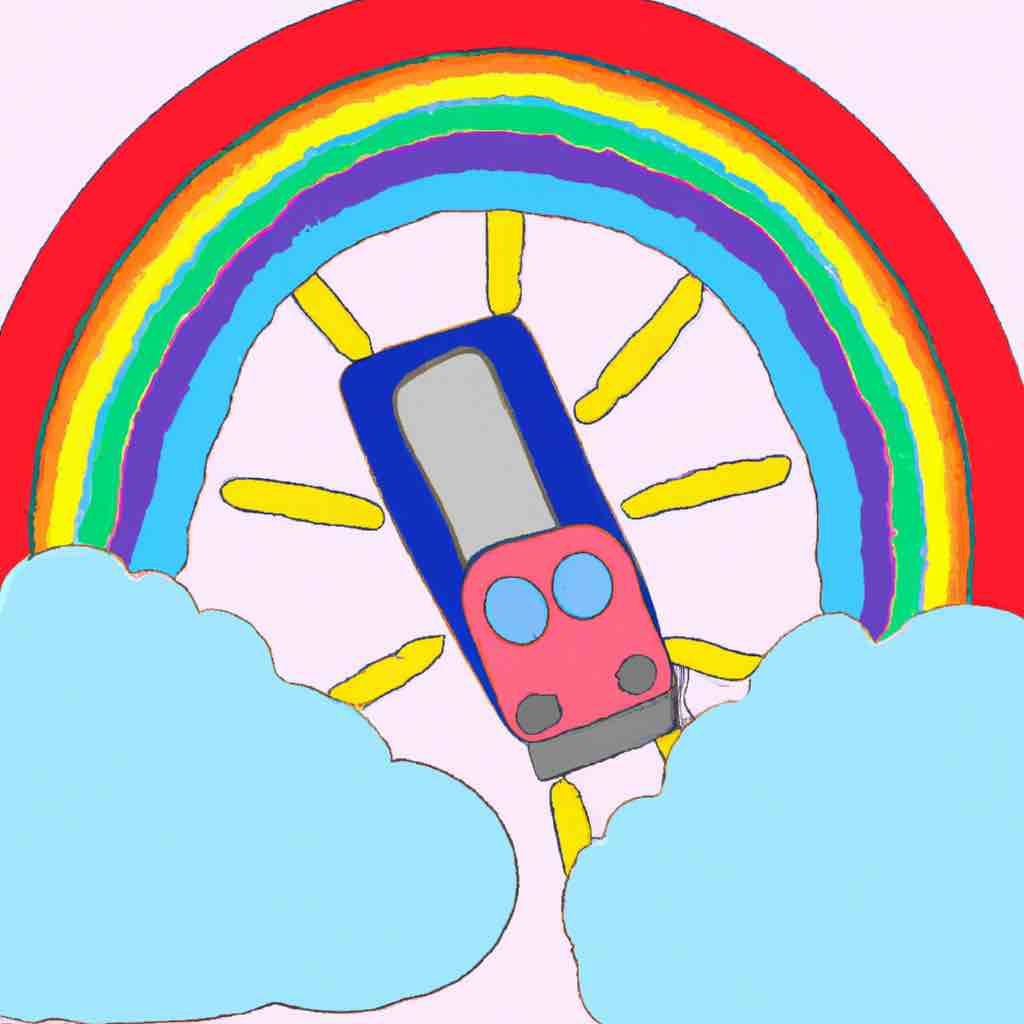} & \includegraphics[width=.4\linewidth]{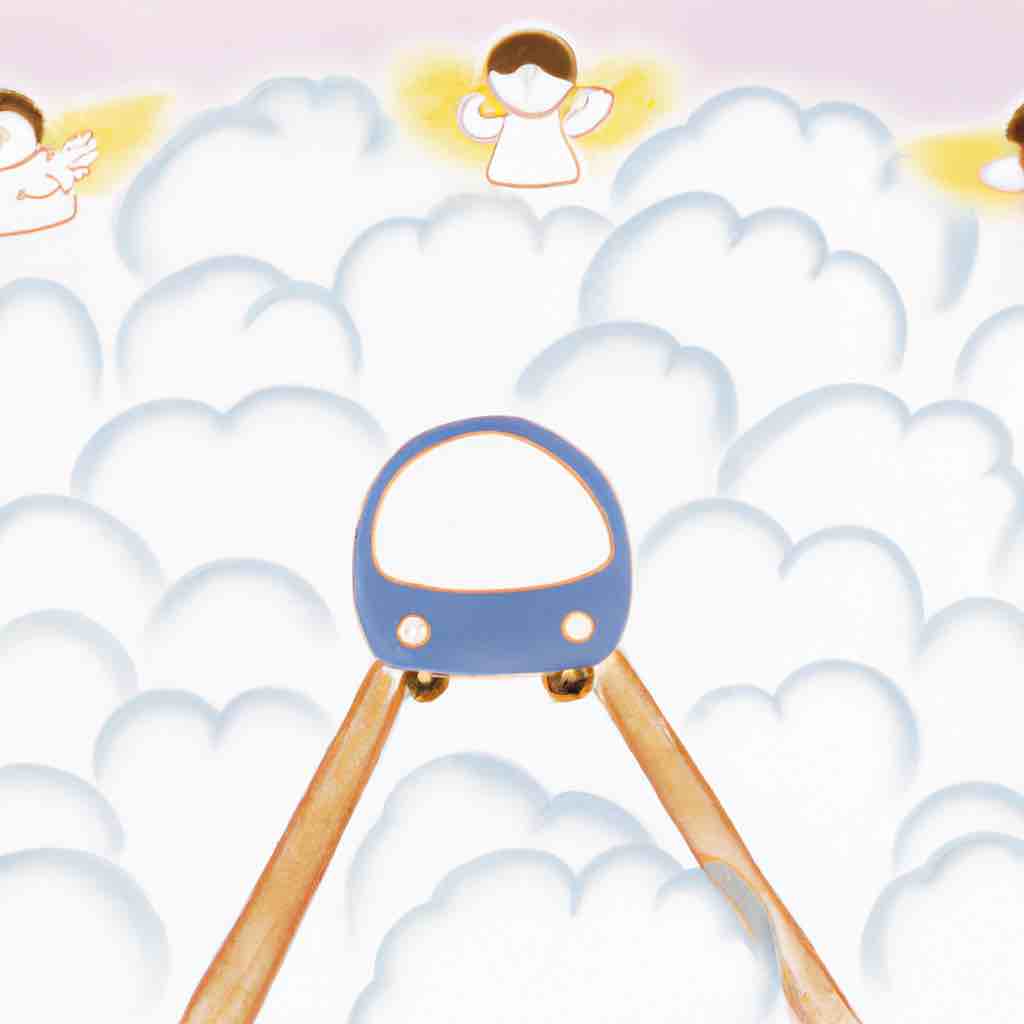} 
\end{tabular}
\caption{Examples of images generated by DALL$\cdot$E 2 prompted with CoT (left) and Completion (right).}
\end{table}

\begin{figure}[!h]
\centering
\includegraphics[width=\linewidth]{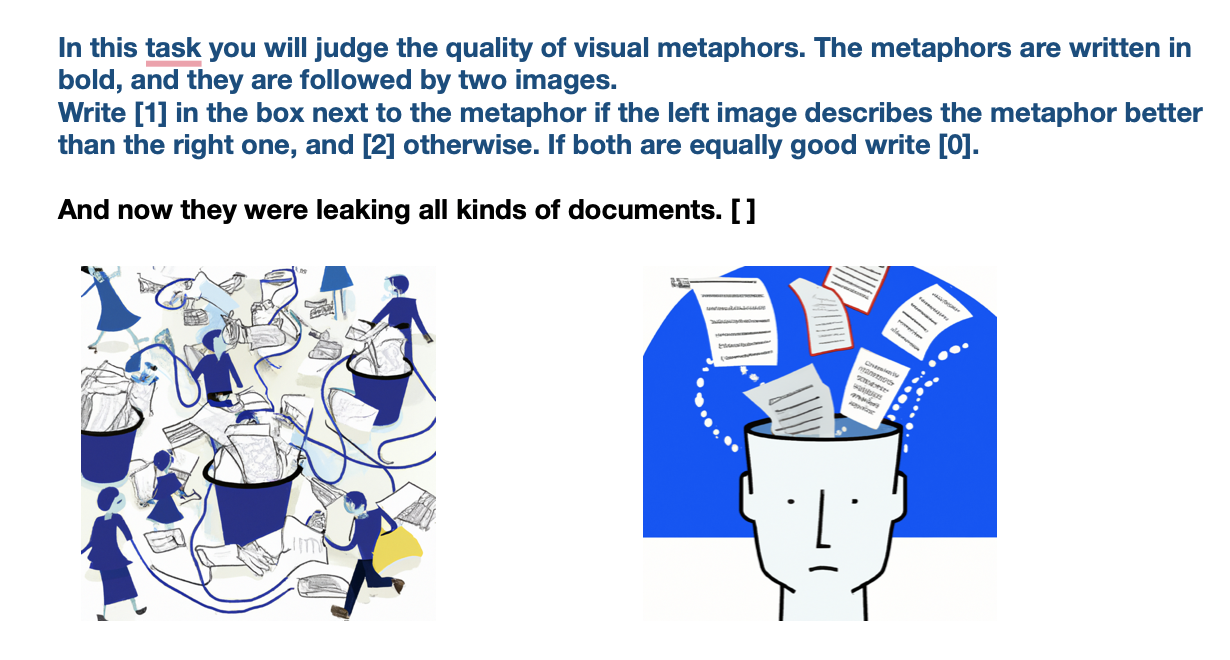}
\caption{Annotators were provided with a list of metaphors along with two images generated by DALL$\cdot$E 2 using our two different prompting methods, CoT and Completion. The order of the images is random.}
\label{fig:prompt_eval_interface}
\end{figure}

\section{Does better prompting lead to better images?} 
Language models are sensitive to prompting \cite{jiang-etal-2020-know}, as are text-to-image diffusion-based models \cite{liu2022design}. We employ CoT prompting to generate visual elaborations of linguistic metaphors using Instruct GPT-3 (davinci-002). The alternative to CoT would be classic Completion prompting, which would require Instruct GPT-3 (davinci-002) to provide visual elaborations for the metaphors without first reasoning about objects and implicit meaning. 

We evaluate whether or not requiring Instruct GPT-3 (davinci-002) to reason about both the included objects and the implicit meaning \textbf{before} providing a visual elaboration improves the quality of the generated visual metaphor, by comparing to Completion prompting where the visual elaboration is directly predicted without the intermediate reasoning steps. For a fair comparison, we require the prompts to be as similar in content as possible, and use the same 5 few-shot examples as for CoT, only removing the intermediate information (objects to be included, implicit meaning) for the Completion prompt.

We verify the hypothesis that CoT improves image quality through a small-scale human evaluation. We consider 50 metaphors for this experiment and generate visual descriptions using the prompt template shown in Table \ref{table:prompt_no_cot}, which replicates the metaphors and visual elaborations in Table \ref{table:prompt} but without the instruction section or the step by step reasoning used in CoT.
The resulting prompts are passed to DALL$\cdot$E 2 to generate images.

We provide 3 annotators with the list of 50 metaphors, as well as the two images that are generated by Instruct GPT-3 (davinci-002) using CoT and Completion prompting without any further post-processing. Figure \ref{fig:prompt_eval_interface} shows the instructions provided to the annotators and an annotation example. To mitigate the subjectivity of the task, which is confirmed by a fair average pairwise Cohen's Kappa score ($\kappa$=0.26), we consider the majority vote selection for each example. Our results show that annotators select 27/50 images that are generated using CoT prompts, 11/50 using Completion prompts, and 12/50 images are judged to be of equal quality regardless of the prompting strategy used. Our results indicate that prompting can significantly improve the quality of the generated images suggesting that future work should investigate ways to further improve the quality of the generated visual metaphors by extracting more detailed specifications from LLMs.

\section{Visual Entailment Data}
\label{app:viz_entailment}
In order to perform the visual entailment task, we require metaphors that are associated with literal hypotheses and their corresponding labels (entailment, contradiction, neutral). \textbf{\texttt{FLUTE}} \cite{chakrabarty2022flute} offers such data without any further processing. For the metaphors in \textbf{\texttt{CrossLing Metaphors}} \cite{tsvetkov-etal-2014-metaphor} and \textbf{\texttt{Metaphor Paraphrases}} \cite{bizzoni-lappin-2018-predicting} we employ \textbf{recasting}, namely \emph{``leveraging existing datasets to create NLI examples''}, \cite{poliak2018collecting} to convert them into textual entailment data. 

The metaphors in \textbf{\texttt{Metaphor Paraphrases}} \cite{bizzoni-lappin-2018-predicting} are each associated with four ranked candidate literal sentence. Each sentence is annotated with a value from 1 to 4, indicating the degree to which the sentence is a paraphrase of the original metaphoric sentence, where 4 stands for exact paraphrase. We consider each sentence and each of the candidate paraphrases as a sentence pair for a textual entailment classification problem and manually annotate them. 

The \textbf{\texttt{CrossLing Metaphors}} \cite{tsvetkov-etal-2014-metaphor} dataset consists of 200 metaphoric English sentences, 200 literal English sentences, and their Russian translations. For the purposes of this study we were only concerned with using the 200 English metaphoric sentences to construct entailment pairs. We manually created three literal hypotheses with corresponding labels (entailment, contradiction, and neutral). 

The data was presented to 3 annotators to verify the quality of the labels. The annotators were presented with both the metaphoric premise and the literal hypothesis, and had to decide whether the hypothesis was entailed, contradicted, or neutral to the statement. The mean pairwise annotator agreement for the labels was .79. The gold label for the data was assigned by majority vote. 

\begin{figure*}[ht]
    \centering
    \includegraphics[width=.94\textwidth]{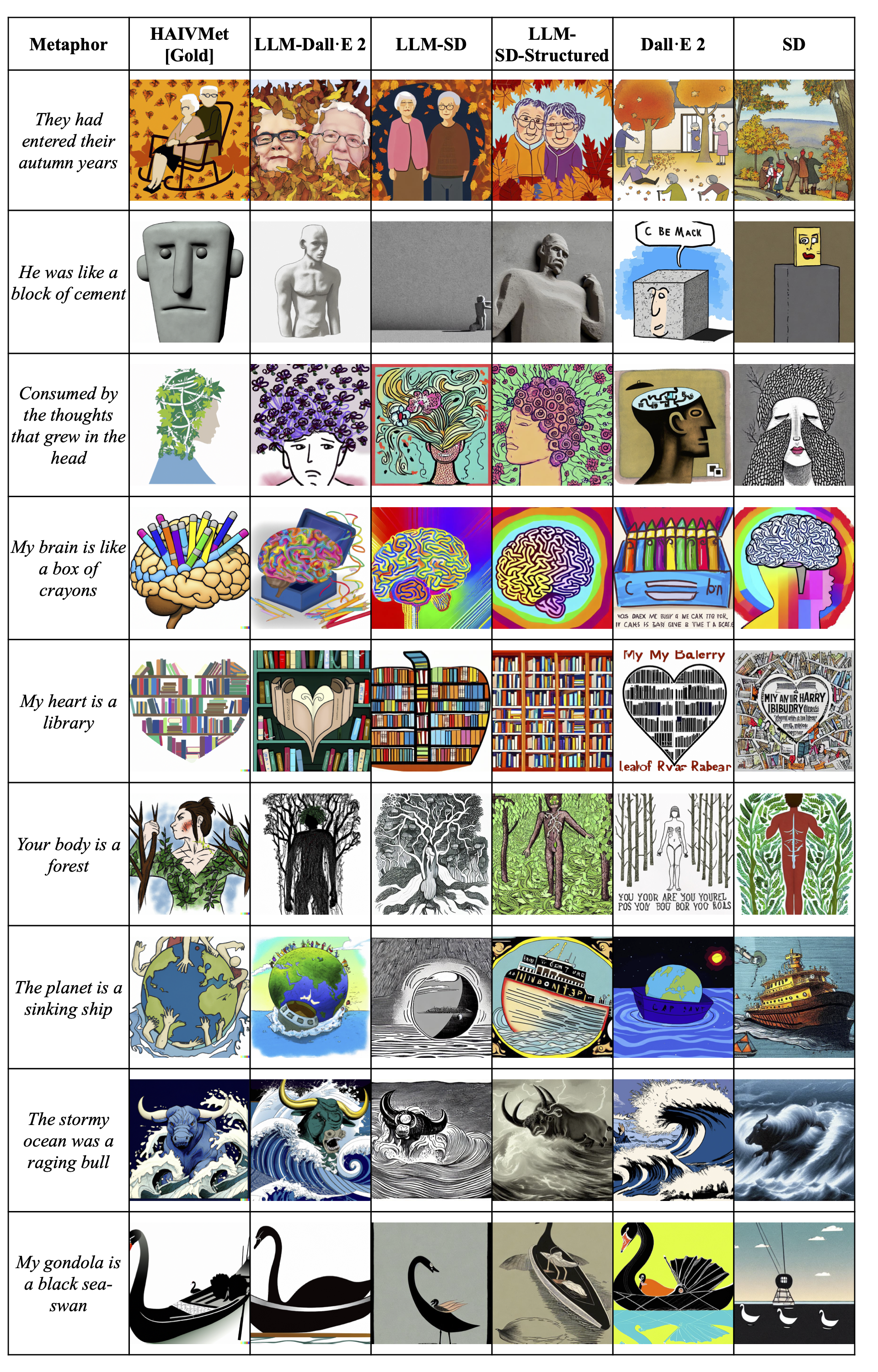}
    \caption{Additional examples of output from the 
    models described in Section \ref{models} for 
    randomly chosen metaphors. \textbf{\texttt{HAIVMet}} is our gold standard. \label{fig:more_qual}}
\end{figure*}

\subsection{Evaluation Interface}
\begin{figure*}[!ht]
\centering
    \subfloat[]{\includegraphics[width=\textwidth]{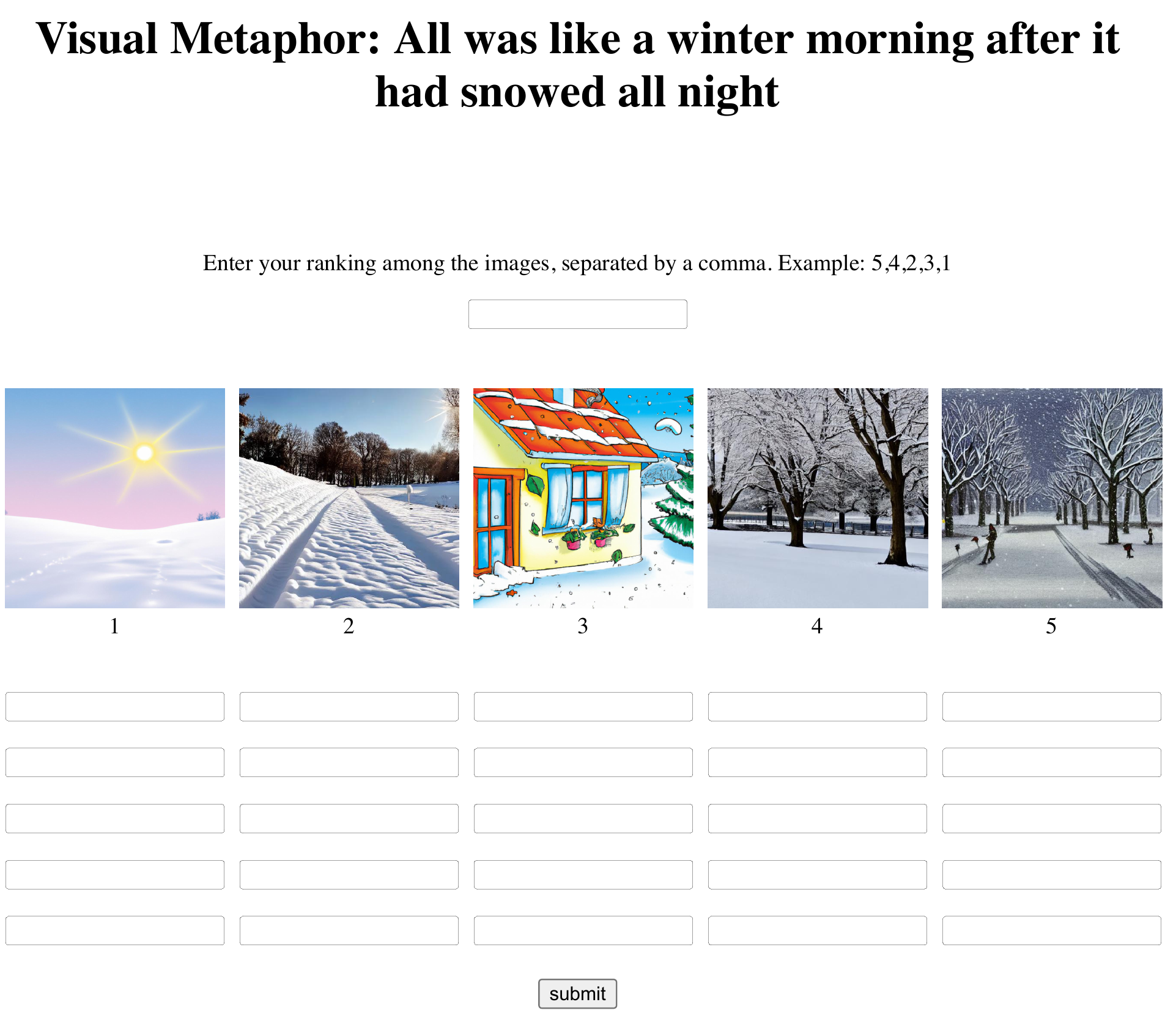}}
    \caption{\label{human1} Evaluation interface for LLM-Diffusion Model collaboration with five systems, as described in Section \ref{sec:eval}.}
    \label{fig:foobar}
\end{figure*}

\begin{figure*}[!ht]
\centering
    \subfloat[]{\includegraphics[width=\textwidth]{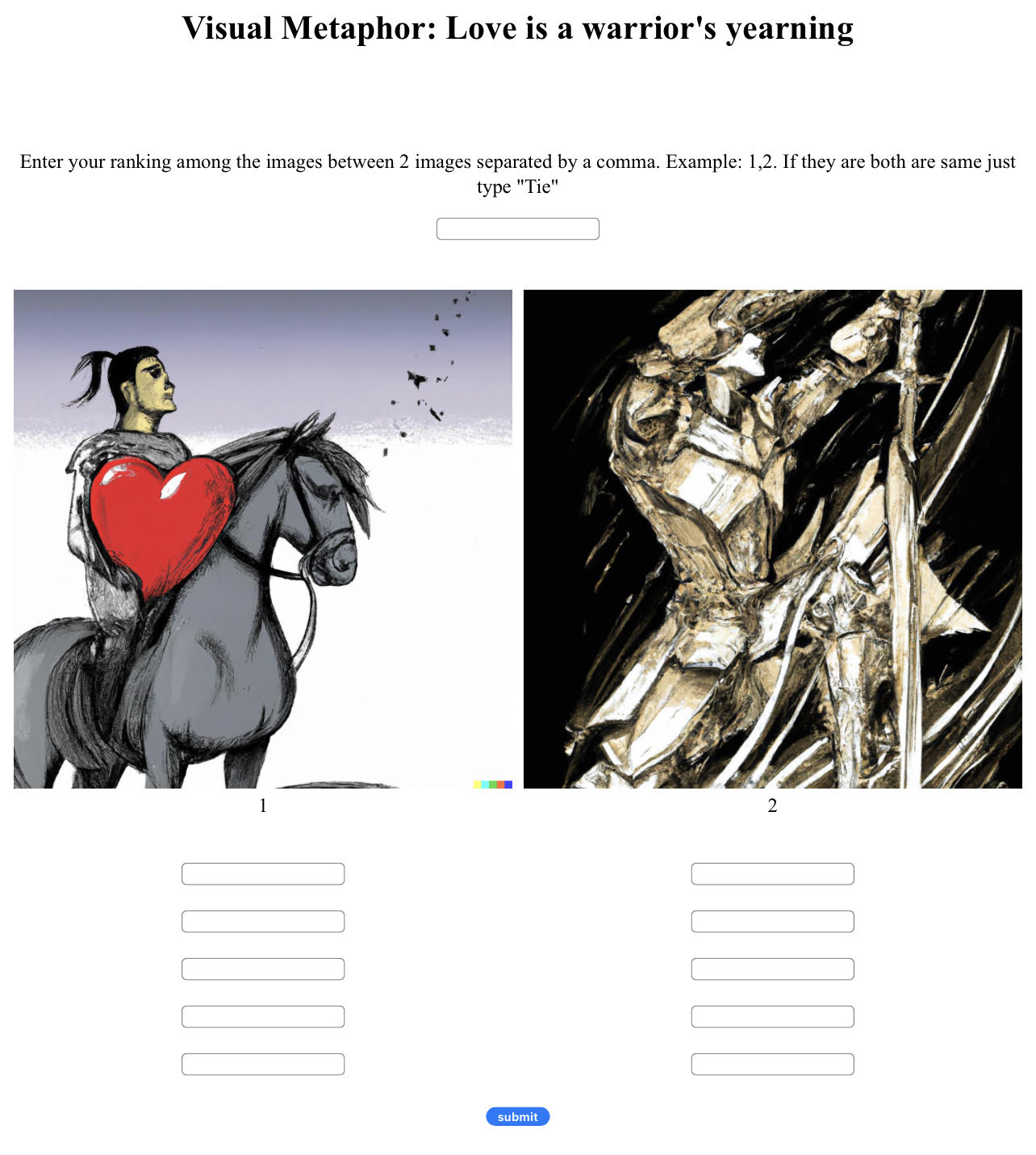}}
    \caption{\label{human2}Evaluation interface for Human AI collaboration with two systems, as described in Section \ref{sec:eval}.}
    \label{fig:foobar2}
\end{figure*}

Figure \ref{human1} and \ref{human2} show the evaluation interface for LLM-Diffusion Model collaboration and Human AI collaboration respectively. For the LLM-Diffusion Model 5 images are presented in randomly shuffled order while for Human AI collaboration 2 images are presented one from LLM-DALLE and the other from \texttt{\textbf{HAIVMet}}.

\end{document}